\documentclass[journal]{IEEEtran}
\usepackage{ifpdf}

\usepackage{booktabs}
\usepackage{multicol}
\usepackage{multirow}
\usepackage{arydshln}
\usepackage{caption}
\usepackage{cite}

\ifCLASSINFOpdf
  \usepackage[pdftex]{graphicx}
\else
\fi
\usepackage{amsmath}
\usepackage{algorithmic}
\begin{document}
%
\title{Adversarial Attacks and Identity Leakage in De-Identification Systems: An Empirical Study}
%
%
%

\author{Felix~Rosberg,
        Cristofer~Englund,
        Eren~Erdal~Aksoy,
        and~Fernando~Alonso-Fernandez}

%
%

\markboth{IEEE Transactions on Biometrics, Behavior, and Identity Science}%
{Rosberg \MakeLowercase{\textit{et al.}}: Adversarial Attacks and Identity Leakage in
De-Identification Systems: An Empirical Study}
%




\maketitle

\begin{abstract}
In this paper, we investigate the impact of adversarial attacks on identity encoders within a realistic de-identification framework. Our experiments show that the transferability of attacks transfers from an external surrogate model to the system model (e.g., CosFace to ArcFace) allows the adversary to cause identity information to leak in a sufficiently sensitive face recognition system. We present experimental evidence and propose strategies to mitigate this vulnerability. Specifically, we show how fine-tuning on adversarial examples helps to mitigate this effect for distortion-based attacks (i.e., snow, fog, etc.), while a simple low-pass filter can attenuate  the effect of adversarial noise without affecting the de-identified images. Our mitigation results in a de-identification system that preserves its functionality while being significantly more robust to adversarial noise.
\end{abstract}

\begin{IEEEkeywords}
De-Identification, Adversarial Attacks, Adversarial Transferability.
\end{IEEEkeywords}

%
\IEEEpeerreviewmaketitle

\section{Introduction}
%
%
%
%

\begin{figure*}[htbp]
\centering
\includegraphics[width=1.0\textwidth]{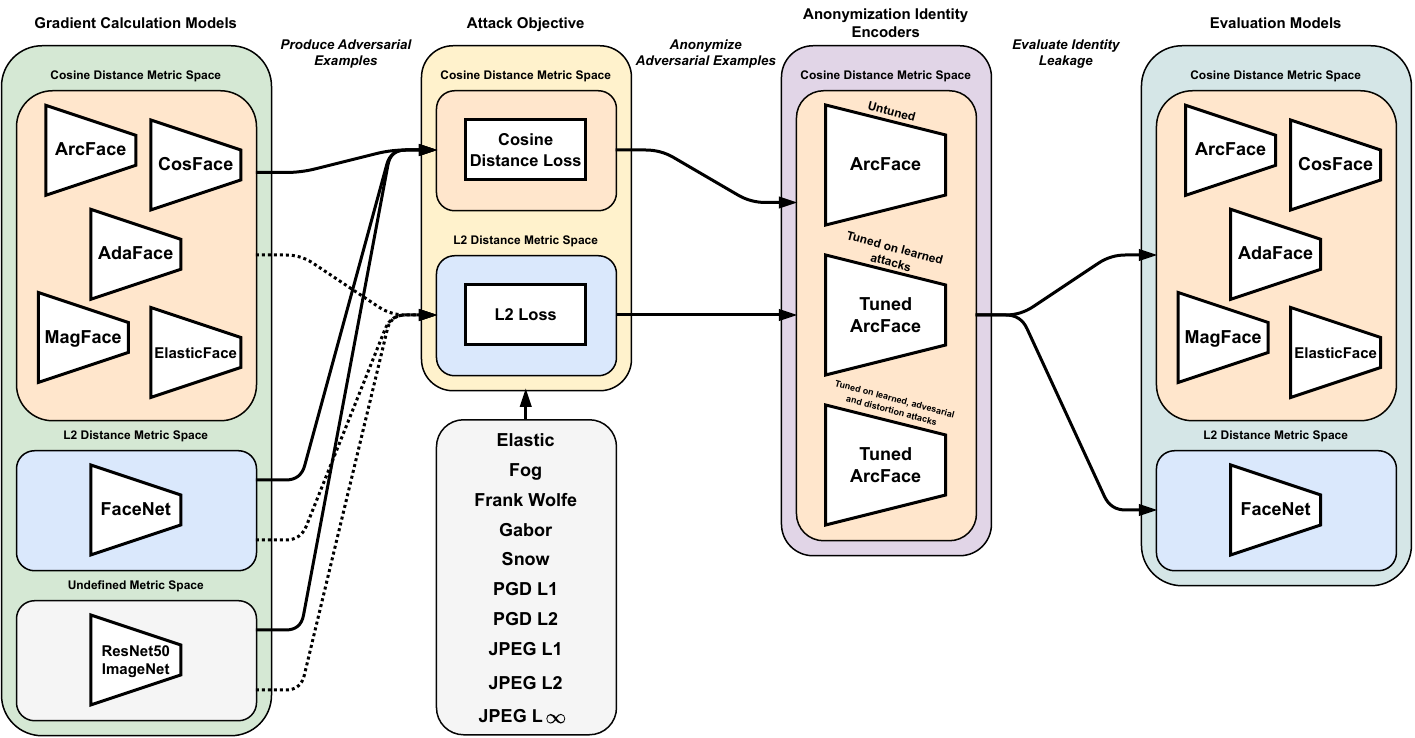}
\caption{
Illustration of the process for creating adversarial examples, anonymization and finally evaluating identity leakage.}
\label{fig:attack_flow}
\end{figure*}

\IEEEPARstart{I}{n} recent years, there has been a notable increase in research efforts at developing realistic de-identification / anonymization methods \cite{FIVA, LIVEDEID, ANONNET, DISENTANGLEDANON, DEEPPRIVACY, DEEPPRIVACY2, towards_privacy_fastzero, MYFACEMYCHOICE, CFANET, ANON_DIFF, DE-ID-VITOMIR, expression}. This trend has arguably been increasing due to the demand of data for deep learning applications in conjunction with data privacy regulations such as the General Data Privacy Regulation (GDPR) \cite{GDPR}. For a large portion of these applications, identity is not of interest, while attributes are. Which in turn has led to numerous advanced methods that aim to obscure the identity while preserving attributes, such as face pose and expression \cite{FIVA, LIVEDEID, ANONNET, DISENTANGLEDANON, MYFACEMYCHOICE, DE-ID-VITOMIR} instead of, for example, blurring. A leading question from this is if there are any vulnerabilities in using generative methods to solve this task. Rosberg et al. \cite{FIVA} showed that target-oriented methods such as \cite{FaceDancer, simswap}, and \cite{FIVA} can be reversed in a black-box attack if enough data is collected from the victim de-identification system, suggesting that target-oriented models leaves a predictable pattern in the de-identified image.

In this work, we extend the subject of security and vulnerabilities in a de-identification system by investigating if the identity encoder used in many of these methods \cite{FIVA, LIVEDEID, DISENTANGLEDANON, ANONNET, MYFACEMYCHOICE, CFANET} can be exploited in such a way that the system leaks identity information. To simulate a certain realism in the attacks themselves, our experiments are based on adversarial transferability \cite{ADV_TRANS_1, ADV_TRANS_2, adversarial_attack_face_rec_transferable, adversarial_attack_defense_suvery}. For our purposes, this means that we assume that an attacker does not have access to the identity encoder used in the de-identification system and has to utilize other identity encoders to produce the adversarial examples. In this work we choose to attack an implementation of FIVA \cite{FIVA} with the help of six identity encoders and one regular pretrained backbone, namely ArcFace \cite{arcface} (Identity encoder used in the victim de-identification system), CosFace \cite{cosface}, AdaFace \cite{adaface}, MagFace \cite{magface}, ElasticFace \cite{elasticface} (Identity encoders that constrain the embeddings to a cosine metric space, similar to ArcFace), FaceNet \cite{FACENET} (Identity encoder that constrain the embeddings to a Euclidean metric space, dissimilar to the other identity encoders), and a ResNet50 \cite{resnet} pretrained on ImageNet \cite{imagenet} where both the model task and embedding space differs to face recognition models completely. Where the main idea is to find answers to, 1: how vulnerable is the system to identity leakage in general towards adversarial attack and 2: how vulnerable is the system to transferable adversarial attacks. Where identity leakage is defined as a successful match in a biometric facial recognition system. For example, our experiments show that attacking using CosFace as a surrogate has a profound effect on ArcFace and thus the de-identification system in terms of identity leakage, while attacking FaceNet does not have the same impact. To clarify further, for differentiable attacks to work, it would either 1) require the attacker to have white-box access to the victim model, or 2) the attacker utilizes surrogate models in a black-box setting, whose adversarial examples transfers to the victim model. The second scenario is mostly of interest here, due to it being an arguable more realistic scenario. This is also important from a robustness perspective, where the real world itself could cause disruptions \cite{adversarial_attack_in_real_world, adversarial_attack_defense_review_real_world_works}. Note, while FIVA \cite{FIVA} is the chosen victim model, our attacks focuses purely on the identity encoders. This means that FIVA acts more as a visualization and measuring tool on the effect on the de-identification task. Which means that any method utilizing identity encoders will experience similar effect as the identity embeddings gets shifted by the adversarial attacks.

We show that the tested adversarial attacks can cause identity leakage. In this work we suggest mitigation by 1) distilling the identity encoder in the de-identification system on adversarial examples and 2) applying a simple preprocessing step in the form of a low-pass filter to dampen high frequency adversarial noise. We empirically demonstrate exhaustive quantitatively results that said adversarial attacks can cause identity leakage and that our proposed mitigation will alleviate this issue significantly. In the context of security, this shows promising results, as the practicality of the attack is being diminished by this.

\section{Related Work}

There have been numerous studies in the domain of adversarial attack \cite{adversarial_attack_defense_on_images_review, adversarial_attack_defense_review_real_world_works, adversarial_attack_on_nn_policy, adversarial_attack_defense_suvery, adversarial_defense_resistance, ADV_TRANS_1, ADV_TRANS_2}, where many studies focus on the image classification task. While research on vulnerabilities for de-identification methods exist \cite{FIVA, REVERSE_ANON}, work on identity leakage due to adversarial attacks remain unexplored. However, there are two main related tasks for this topic. 1) Adversarial attacks on facial recognition models \cite{adversarial_attack_face_rec_1, adversarial_attack_face_rec_study, adversarial_attack_face_rec_transferable, AdvHat, AdvMakeup, AdvSticker} and, 2) adversarial attacks on image translation models \cite{adversarial_i2i_nullifying, adversarial_i2i_segmentation, adversarial_i2i_deepfake_1, adversarial_i2i_deepfake_2, adversarial_i2i_deepfake_3, adversarial_i2i_faceswap}. For image translation models, it includes adversarial attacks on deepfake and face swapping methods \cite{adversarial_i2i_deepfake_1, adversarial_i2i_deepfake_2, adversarial_i2i_deepfake_3, adversarial_i2i_faceswap}, which is highly related to the realistic de-identification task, as it can be seen as a special case of face swapping.

In regard to adversarial attacks on facial recognition models, there is work that demonstrates real world practicality. Where one tested approach is wearable perturbations \cite{AdvHat, AdvSticker, AdvMakeup, adversarial_attack_defense_review_real_world_works, adversarial_attack_in_real_world}. Particularly, \cite{AdvHat, AdvSticker} showed that a wearable sticker on the face can cause the identity encoder to misjudge the identity. In this work we focus on adversarial attacks against facial recognition models due to many methods for de-identification rely on target-oriented models \cite{FIVA, LIVEDEID, DISENTANGLEDANON, ANONNET, MYFACEMYCHOICE, CFANET} which uses said facial recognition models as a prior for controlling the synthetic identity. If there is a possibility of causing this part of the de-identification system to misjudge the identity information, it may cause the true identity to leak.

\section{Method}
\label{s:method}


\subsection{Gradient Calculation Models}
\label{ss:gradient_calc_models}
\vspace{0.8mm}\noindent We have selected seven models to compute gradients for adversarial attacks, comprising six state-of-the-art identity encoders employed in facial recognition systems: ArcFace \cite{arcface}, CosFace \cite{cosface}, AdaFace \cite{adaface}, MagFace \cite{magface}, ElasticFace \cite{elasticface}, and FaceNet \cite{FACENET}, alongside a ResNet50 \cite{resnet} pretrained on ImageNet \cite{imagenet} (see Figure \ref{fig:attack_flow}, first column). Our selection rationale encompasses several considerations. ArcFace is chosen because it mirrors the identity encoder utilized in the victim de-identification system (white-box attack). CosFace, AdaFace, MagFace and ElasticFace are included due to its training methodology, which constrains embeddings within a cosine distance metric space, same as ArcFace. FaceNet, in contrast, operates in an L2 distance metric space, different from ArcFace offering a diverse perspective. Finally, ResNet50 pretrained on ImageNet is incorporated due to its distinct 'undefined' embedding space and disparate training data. To clarify, cosine distance metric space refers to encoders that constrain the latent representation into a hyper-sphere during training. L2 distance metric space refers to encoders that constrain the latent representation into a Euclidean space during training. Undefined metric space refers to encoders trained on separate tasks, i.e., classification. Note that the resulting latent representation could still have a L2 or angular nature. This diverse selection is motivated by the concept of transferability, where adversarial perturbations generated using one model may impact another \cite{ADV_TRANS_1, ADV_TRANS_2}. FaceNet and the ResNet50 pretrained on ImageNet are included to investigate if models with a differently constrained feature space are viable for transferability for attacking the de-identification systems.

\subsection{Attack Objectives}
\label{ss:attack_objectives}
\vspace{0.8mm}\noindent Our choice of loss functions is similarly motivated by the metric spaces pertinent to each model (see Figure \ref{fig:attack_flow}, second column). Given that the victim system employs ArcFace and the current best practice of cosine distance-constrained embedding spaces, we utilize cosine distance as the primary metric. Consequently, we generate adversarial examples using cosine distance for all gradient calculation models, ensuring compatibility with the target system. As FaceNet, which inherently operates within an L2 distance metric space, we adopt L2 loss to maintain fidelity with its original embedding space. However, we perform experiments using both loss functions for each surrogate model to see if there is any importance of matching the victim model's embedding space.

\subsection{De-Identification Identity Encoders}
\label{ss:fiva_id_encoders}
\begin{figure}[htbp]
\centering
\includegraphics[width=0.5\textwidth]{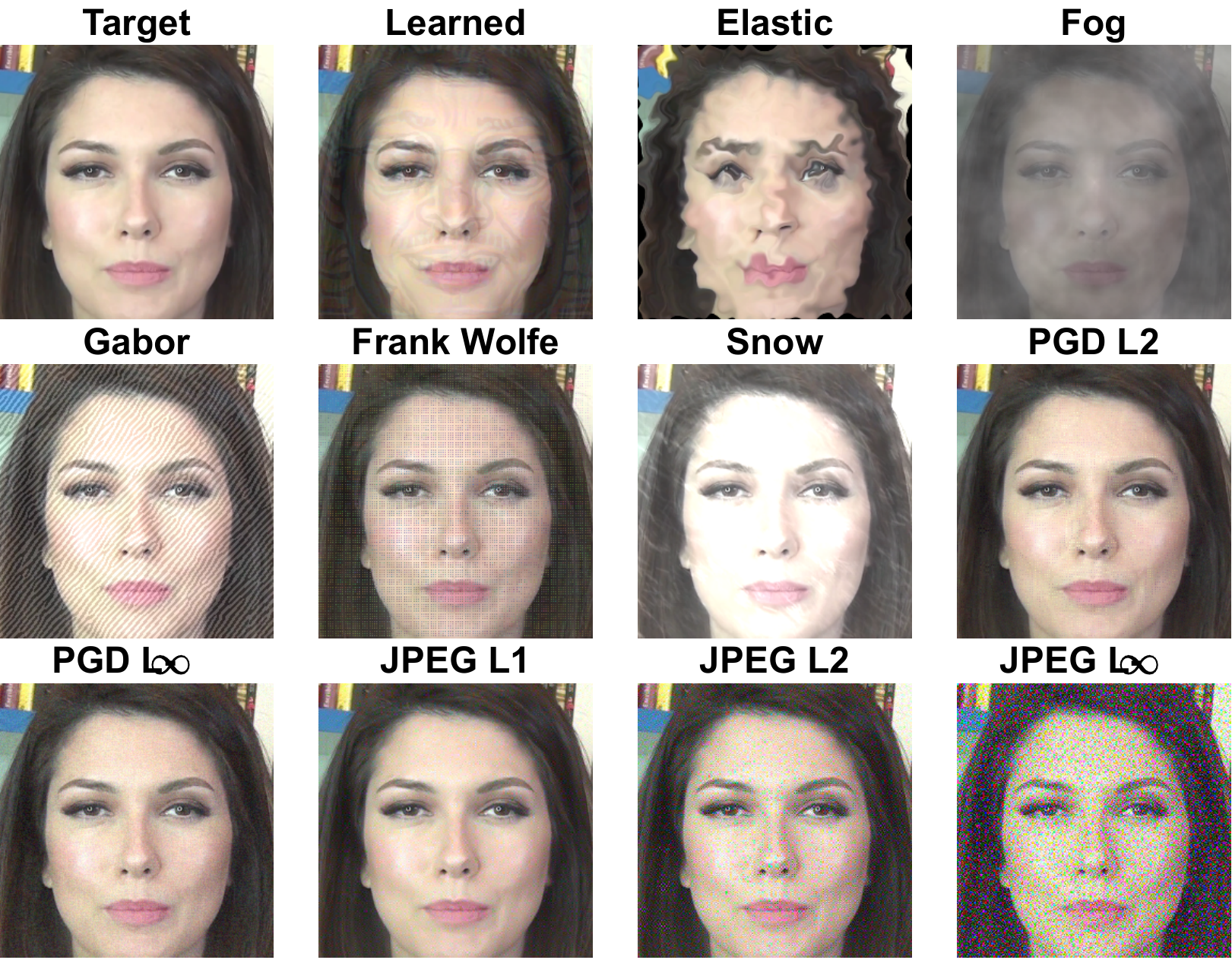}
\caption{
Illustration of the adversarial noises used.}
\label{fig:attack_examples}
\end{figure}

\vspace{0.8mm}\noindent After adversarial example generation, we subject the images to de-identification using the victim system, instantiated as an implementation of FIVA \cite{FIVA}. Our experiments evaluate three variants of ArcFace: the original version utilized by FIVA, ArcFace fine-tuned against learned perturbations \cite{ReFace} (see Figure \ref{fig:attack_flow}, Figure \ref{fig:attack_examples}: Learned and Figure \ref{fig:attack_framwork}b and c), and ArcFace fine-tuned against a combination of adversarial attacks and distortion-based attacks \cite{PGD, frankwolfe, OpenAIRobust} (refer to Figure \ref{fig:attack_framwork}b).

\subsection{Adversarial Attacks}
\label{ss:adversarial_attacks}
\vspace{0.8mm}\noindent We opt for a diverse set of adversarial attacks to try; specifically, we utilize the framework introduced in \cite{OpenAIRobust}. To be even more specific, we use the attacks provided by the advex-uar repository\footnote{https://github.com/ddkang/advex-uar}. Originally, the code is targeted towards classification models; however, the identity encoders used in our experiments are not using a classification head for practical applications. They produce an embedding that can be used to compare with other embeddings through a distance metric such as L2 and cosine distance. Thus, we adapt the objective to the loss functions mentioned in Section \ref{ss:attack_objectives} and shown in Figure \ref{fig:attack_flow}, column 2. The attacks included are: an elastic transform, fog, Gabor, Frank Wolfe attack, snow, JPEG compression, and projected gradient descent (see Figure \ref{fig:attack_examples}). The framework lets the user define the strength $\epsilon$ for each attack, which has an arbitrary range for different attacks. We make a qualitative decision for the strength for each attack that is based on the resulting loss from calculating the gradients when producing an attack and the visual effect on the images (see Table \ref{t:attack_strength}). We do this as the amount of experiments that need to be run would grow exponentially across the number of attacks, number of evaluation models, and number of surrogate models. Learned attack, as introduced in \cite{ReFace}, is only used to finetune the victim de-identification system's identity encoder ArcFace (see Figure \ref{fig:attack_framwork}).

\subsection{Evaluation}
\label{ss:evaluation}
\begin{figure*}[htbp]
\centering
\includegraphics[width=\textwidth]{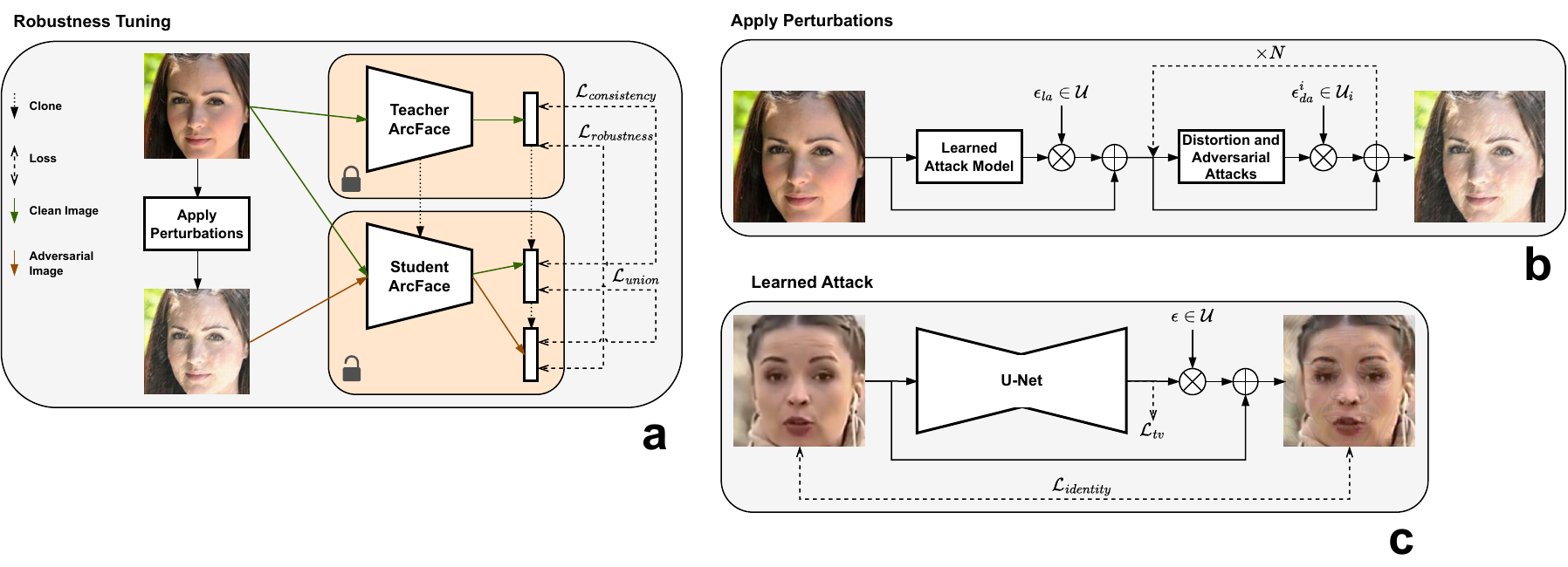}
\caption{
Illustration of a) robustness fine-tuning the victim systems' identity encoder (ArcFace), b) How we apply a mixture of different perturbations (we set $N$ to 3), with varying strength ($\epsilon_{la}$, $\epsilon_{da}$), and c) training of the learned perturbation model.}
\label{fig:attack_framwork}
\end{figure*}

\vspace{0.8mm}\noindent The framework provided by \cite{OpenAIRobust} introduces a robustness metric based on the combined accuracy of each attack on the victim model. We chose to omit this in our work for a couple of reasons: 1) We attack an identity encoder that provides embeddings and not classification logits, and 2) Even if the attack is targeted against the identity encoders, our actual target is the de-identification system as a whole. This means that we opt for evaluating the produced de-identified faces. We focus on identity retrieval \cite{FIVA, LIVEDEID, DEEPPRIVACY, ANONNET, CFANET}, for three different false acceptance rate (FAR) values: $10^{-3}$, $10^{-4}$ and $10^{-5}$. We evaluate identity retrieval with the six identity encoders as shown in Figure \ref{fig:attack_flow} column 4, to address the investigation of transferability further. Attacking ArcFace and then only evaluating with ArcFace makes little sense for the same reason. The evaluation is done for each configuration of attacks, identity encoders, and de-identification systems. We use FaceForensic++ \cite{faceforensics++} as the chosen evaluation dataset. Due to a number of combinations between gradient calculation models, evaluation models, different attacks, different FAR thresholds, and different loss functions, the results are aggregated. See Supplementary Materials for detailed results for each attack, evaluation model, and surrogate model combination.
\\
\\
As described below in Section \ref{ss:robustness_tuning}, shown in Figure \ref{fig:attack_flow} column 3 and demonstrated in Figure \ref{fig:attack_framwork}a we fine tune ArcFace with adversarial examples. Evaluation of the resulting models is done to investigate if there is any significant reduction in performance. We calculate the false rejection rate (FRR) for the FAR values $10^{-3}$, $10^{-4}$ and $10^{-5}$, and the accuracy for the same FAR values. We use FaceForensic++ \cite{faceforensics++} and Labeled Faces in the Wild \cite{LFW} (LFW) as the chosen evaluation datasets.

\subsection{Robustness Tuning}
\label{ss:robustness_tuning}
\vspace{0.8mm}\noindent The expected effect of the aforementioned adversarial attacks is that the de-identification system may leak the original identity. This is also shown in Table \ref{t:evaluation_id_baseline}. To alleviate the effect, we fine-tune the de-identification systems' identity encoder to be robust against adversarial examples. We do this through a knowledge distillation framework \cite{DISTILLATION_SURVEY} where the student model is fed both clean and adversarial images (see Figure \ref{fig:attack_framwork}a). The student model is initialized as a clone of the original pretrained ArcFace model. Furthermore, we clone the final linear layer and batch normalization layer, so the student model has two of each. One is responsible for matching the embedding of a clean image when given an adversarial example, while the other is responsible for maintaining the information when given clean images. We keep the fine-tuning step simple by optimizing the minimization of the cosine distance between the teacher and the student. Let $z_{tc}$, $z_{sc}$, $z_{sa}$ be clean from the teacher, the clean embeddings from the student (from the first embedding layers), and the adversarial embeddings from the student (from the cloned embedding layers), the loss function is:

\begin{equation}
\begin{aligned}
    \mathcal{L} = \mathcal{L}_{robustness} * \lambda_{robustness} + \\
                  \mathcal{L}_{consistency} * \lambda_{consistency} + \\
                  \mathcal{L}_{union} * \lambda_{union} ~,
    \label{eq:robustness_loss}
\end{aligned}
\end{equation}

\vspace{0.8mm}\noindent where $\mathcal{L}_{robustness}$ is the cosine distance between $z_{tc}$ and $z_{sa}$, $\mathcal{L}_{consistency}$ is the cosine distance between $z_{tc}$ and $z_{sc}$, and $\mathcal{L}_{union}$ is the cosine distance between $z_{sc}$ and $z_{sa}$. $\lambda_{robustness}, \lambda_{consistency}, \lambda_{union}$ are the loss weighting coefficients.

\subsection{Explainable Inspection}
\label{ss:xai}
\vspace{0.8mm}\noindent To visualize the effect of the adversarial attacks, we use a DeepFaceDecoder introduced in \cite{DeepFaceDecoder}. DeepFaceDecoder is trained to reconstruct the face purely from the identity embedding. This let us visually inspect the effect the attacks have on embeddings.

\section{Results}
\label{s:results}


\begin{table}[htbp]
\caption{Strength of noise chosen and number of gradient calculation iterations $N$ for each attack in accordance with the implementation in the advex-uar repository and the strength of perturbation produced by the learned attack. }
\label{t:attack_strength}
\begin{center}
\begin{tabular}{cccc}
\toprule
Attack Type & $\epsilon$ & $N$ & Category\\
\hline
Elastic & 8 & 50 & Distortion\\
Fog & 600 & 50 & Distortion\\
Gabor & 50 & 50 & Distortion\\
Frank Wolfe & 13 & 50 & Adversarial noise\\
Snow & 0.125 & 50 & Distortion\\
PGD L2 & 2400 & 50 & Adversarial noise\\
PGD L-$\infty$ & 16 & 50 & Adversarial noise\\
JPEG L1 & 131072 & 50 & Distortion\\
JPEG L2 & 128 & 50 & Distortion\\
JPEG L-$\infty$ & 2 & 50 & Distortion\\
\bottomrule
\end{tabular}
\end{center}
\end{table}

\begin{figure}[htbp]
\centering
\includegraphics[width=0.5\textwidth]{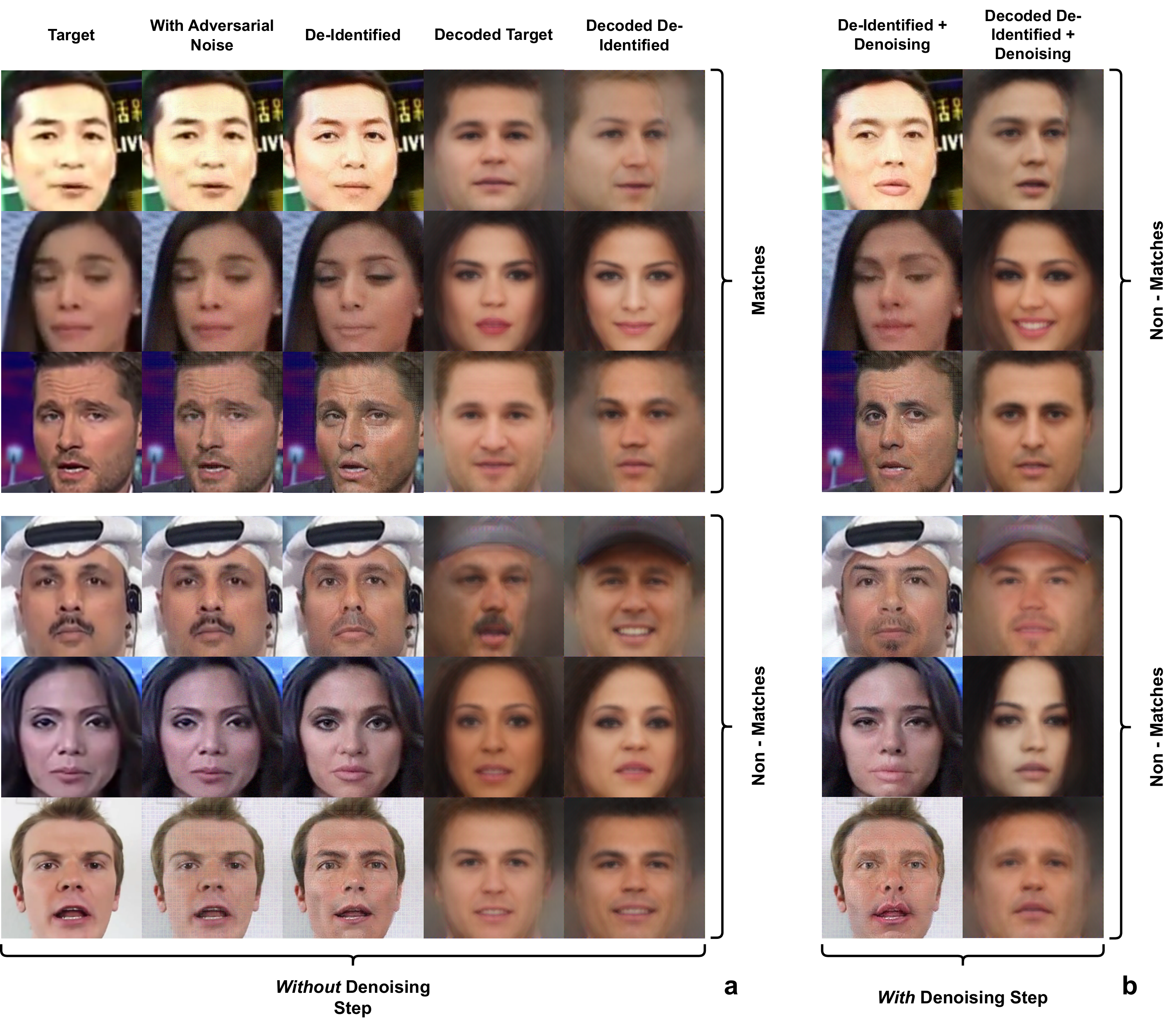}
\caption{
a) Examples of when an attack causes a match during identity retrieval and when the attack fails and does not match with the correct identity. We also show the decoded ArcFace embeddings of the target and the de-identified attacked target to visualize similarities. b) De-identified attacked target and corresponding decoded embeddings after applying a Gaussian blur filter to denoise the input into the identity encoder in the victim de-identification system. These particular examples are for a FAR value of $10^{-4}$.}
\label{fig:match_non_match}
\end{figure}

\begin{table*}[htbp]
\caption{Quantitative experiments on FaceForensics++~\cite{faceforensics++}. We evaluate successful identity retrieval (lower is better in terms of de-identification) after de-identifying the adversarial images using FIVA \cite{FIVA}. We show an average aggregation for each attack. Furthermore, we also show an average aggregation for each evaluation model per FAR threshold. To the right you also find the performance of the de-identification system without any adversarial attacks.}
\label{t:evaluation_id_baseline}
\begin{center}
\resizebox{1.0\textwidth}{!}{
\begin{tabular}{c|cccccc:c|cccccc:c|c}
\toprule
&\multicolumn{14}{c}{Gradient Calculation Models} \\
&\multicolumn{7}{c}{Cosine Distance Objective} &\multicolumn{7}{c}{Euclidean Distance Objective}\\
Evaluation Models            & ResNet & AdaFace & MagFace & ElasticFace & CosFace & FaceNet & ArcFace & ResNet & AdaFace & MagFace & ElasticFace & CosFace & FaceNet & ArcFace & No Attack \\
\cline{1-16}
ArcFace FAR $10^{-3}$ ($<0.76$)        & 0.0026 & 0.0647 & 0.0305   & 0.0829      & 0.0780  & 0.0046 & 0.2187 & 0.0025 & 0.0640 & 0.0297 & 0.0829 & 0.0771 & 0.0042 & 0.2169 & 0.0000 \\
ArcFace FAR $10^{-4}$ ($<0.69$)        & 0.0006 & 0.0343 & 0.0130   & 0.0481      & 0.0424  & 0.0014 & 0.1738 & 0.0005 & 0.0329 & 0.0120 & 0.0472 & 0.0424 & 0.0011 & 0.1725 & 0.0000 \\
ArcFace FAR $10^{-5}$ ($<0.62$)        & 0.0001 & 0.0075 & 0.0020   & 0.0121      & 0.0104  & 0.0001 & 0.0866 & 0.0000 & 0.0066 & 0.0017 & 0.0113 & 0.0105 & 0.0001 & 0.0853 & 0.0000 \\
\cdashline{1-16}
AdaFace FAR $10^{-3}$ ($<0.80$)        & 0.0078 & 0.0894 & 0.0450   & 0.1146      & 0.1094  & 0.0094 & 0.2668 & 0.0074 & 0.0880 & 0.0443 & 0.1119 & 0.1072 & 0.0089 & 0.2669 & 0.0000 \\
AdaFace FAR $10^{-4}$ ($<0.75$)        & 0.0034 & 0.0609 & 0.0262   & 0.0821      & 0.0751  & 0.0040 & 0.2251 & 0.0029 & 0.0600 & 0.0250 & 0.0807 & 0.0735 & 0.0040 & 0.2240 & 0.0000 \\
AdaFace FAR $10^{-5}$ ($<0.70$)        & 0.0004 & 0.0242 & 0.0071   & 0.0348      & 0.0309  & 0.0008 & 0.1416 & 0.0005 & 0.0230 & 0.0068 & 0.0350 & 0.0301 & 0.0007 & 0.1412 & 0.0000 \\

MagFace FAR $10^{-3}$ ($<0.31$)        & 0.0625 & 0.0651 & 0.0541   & 0.0672      & 0.0682  & 0.0594 & 0.0797 & 0.0632 & 0.0651 & 0.0540 & 0.0672 & 0.0682 & 0.0579 & 0.0788 & 0.0018 \\
MagFace FAR $10^{-4}$ ($<0.18$)        & 0.0298 & 0.0285 & 0.0235   & 0.0291      & 0.0304  & 0.0277 & 0.0318 & 0.0299 & 0.0285 & 0.0233 & 0.0293 & 0.0302 & 0.0273 & 0.0314 & 0.0016 \\
MagFace FAR $10^{-5}$ ($<0.09$)        & 0.0098 & 0.0095 & 0.0077   & 0.0099      & 0.0100  & 0.0094 & 0.0113 & 0.0098 & 0.0096 & 0.0074 & 0.0101 & 0.0104 & 0.0092 & 0.0112 & 0.0000 \\

ElasticFace FAR $10^{-3}$ ($<0.80$)    & 0.0089 & 0.0874 & 0.0445   & 0.1062      & 0.1052  & 0.0091 & 0.2538 & 0.0082 & 0.0870 & 0.0427 & 0.1052 & 0.1035 & 0.0092 & 0.2539 & 0.0000 \\
ElasticFace FAR $10^{-4}$ ($<0.76$)    & 0.0049 & 0.0683 & 0.0312   & 0.0881      & 0.0829  & 0.0048 & 0.2302 & 0.0047 & 0.0683 & 0.0303 & 0.0864 & 0.0825 & 0.0050 & 0.2296 & 0.0000 \\
ElasticFace FAR $10^{-5}$ ($<0.70$)    & 0.0007 & 0.0245 & 0.0079   & 0.0344      & 0.0301  & 0.0008 & 0.1434 & 0.0008 & 0.0231 & 0.0079 & 0.0336 & 0.0294 & 0.0007 & 0.1418 & 0.0000 \\

CosFace FAR $10^{-3}$ ($<0.78$)        & 0.0077 & 0.0925 & 0.0477   & 0.1137      & 0.0955  & 0.0091 & 0.2659 & 0.0074 & 0.0927 & 0.0470 & 0.1115 & 0.0950 & 0.0089 & 0.2668 & 0.0001 \\
CosFace FAR $10^{-4}$ ($<0.72$)        & 0.0029 & 0.0581 & 0.0247   & 0.0756      & 0.0615  & 0.0033 & 0.2170 & 0.0003 & 0.0570 & 0.0236 & 0.0745 & 0.0614 & 0.0034 & 0.2172 & 0.0000 \\
CosFace FAR $10^{-5}$ ($<0.60$)        & 0.0000 & 0.0038 & 0.0008   & 0.0059      & 0.0046  & 0.0000 & 0.0522 & 0.0000 & 0.0033 & 0.0007 & 0.0048 & 0.0048 & 0.0001 & 0.0498 & 0.0000 \\

FaceNet FAR $10^{-3}$ ($<0.93$)        & 0.0208 & 0.0421 & 0.0329   & 0.0508      & 0.0531  & 0.0128 & 0.1204 & 0.0185 & 0.0414 & 0.0317 & 0.0500 & 0.0531 & 0.0120 & 0.1193 & 0.0084 \\
FaceNet FAR $10^{-4}$ ($<0.71$)        & 0.0003 & 0.0010 & 0.0007   & 0.0013      & 0.0017  & 0.0002 & 0.0047 & 0.0003 & 0.0011 & 0.0008 & 0.0018 & 0.0016 & 0.0002 & 0.0057 & 0.0001 \\
FaceNet FAR $10^{-5}$ ($<0.59$)        & 0.0000 & 0.0000 & 0.0000   & 0.0001      & 0.0001  & 0.0000 & 0.0004 & 0.0000 & 0.0000 & 0.0000 & 0.0001 & 0.0001 & 0.0000 & 0.0002 & 0.0000 \\
\cline{1-16}
Average FAR $10^{-3}$        & 0.0183 & 0.0735 & 0.0425   & \textbf{0.0892}      & 0.0849  & 0.0174 & \emph{0.2009} & 0.0178 & 0.0730 & 0.0416 & \textbf{0.0881} & 0.0840 & 0.0169 & \emph{0.2004} & 0.0017 \\
Average FAR $10^{-4}$        & 0.0070 & 0.0418 & 0.0199   & \textbf{0.0541}      & 0.0490  & 0.0069 & \emph{0.1471} & 0.0069 & 0.0413 & 0.0192 & \textbf{0.0533} & 0.0486 & 0.0068 & \emph{0.1467} & 0.0003 \\
Average FAR $10^{-5}$        & 0.0018 & 0.0116 & 0.0043   & \textbf{0.0162}      & 0.0144  & 0.0019 & \emph{0.0726} & 0.0018 & 0.0109 & 0.0041 & \textbf{0.0158} & 0.0142 & 0.0018 & \emph{0.0716} & 0.0000 \\

\bottomrule

\end{tabular}
}
\end{center}
\end{table*}

\begin{table*}[htbp]
\caption{Quantitative experiments on FaceForensics++~\cite{faceforensics++} after tuning the victim system's identity encoder (ArcFace) on learned adversarial examples \cite{ReFace}. We evaluate successful identity retrieval (lower is better in terms of de-identification) after de-identifying the adversarial images using FIVA \cite{FIVA}. We show an average aggregation for each attack. Furthermore, we also show an average aggregation for each evaluation model per FAR threshold.}
\label{t:evaluation_id_tuned_system}
\begin{center}
\resizebox{1.0\textwidth}{!}{
\begin{tabular}{c|cccccc:c|cccccc:c|c}
\toprule
&\multicolumn{14}{c}{Gradient Calculation Models} \\
&\multicolumn{7}{c}{Cosine Distance Objective} &\multicolumn{7}{c}{Euclidean Distance Objective}\\
Evaluation Models            & ResNet & AdaFace & MagFace & ElasticFace & CosFace & FaceNet & ArcFace & ResNet & AdaFace & MagFace & ElasticFace & CosFace & FaceNet & ArcFace & No Attack \\
\cline{1-16}
ArcFace FAR $10^{-3}$ ($<0.76$)        & 0.0043 & 0.0531 & 0.0258 & 0.0568 & 0.0489 & 0.0070 & 0.1133 & 0.0040 & 0.0523 & 0.0267 & 0.0562 & 0.0483 & 0.0065 & 0.1145 & 0.0000 \\
ArcFace FAR $10^{-4}$ ($<0.69$)        & 0.0008 & 0.0240 & 0.0095 & 0.0269 & 0.0177 & 0.0017 & 0.0733 & 0.0005 & 0.0235 & 0.0092 & 0.0267 & 0.0174 & 0.0017 & 0.0735 & 0.0000 \\
ArcFace FAR $10^{-5}$ ($<0.62$)        & 0.0001 & 0.0040 & 0.0015 & 0.0046 & 0.0021 & 0.0001 & 0.0244 & 0.0000 & 0.0038 & 0.0010 & 0.0048 & 0.0021 & 0.0002 & 0.0232 & 0.0000 \\
\cdashline{1-16}
AdaFace FAR $10^{-3}$ ($<0.80$)        & 0.0105 & 0.0735 & 0.0397 & 0.0844 & 0.0721 & 0.0116 & 0.1588 & 0.0101 & 0.0709 & 0.0390 & 0.0837 & 0.0730 & 0.0110 & 0.1622 & 0.0000 \\
AdaFace FAR $10^{-4}$ ($<0.75$)        & 0.0038 & 0.0453 & 0.0201 & 0.0538 & 0.0383 & 0.0046 & 0.1150 & 0.0036 & 0.0448 & 0.0201 & 0.0539 & 0.0381 & 0.0042 & 0.1186 & 0.0000 \\
AdaFace FAR $10^{-5}$ ($<0.70$)        & 0.0003 & 0.0158 & 0.0048 & 0.0183 & 0.0098 & 0.0007 & 0.0527 & 0.0003 & 0.0153 & 0.0047 & 0.0186 & 0.0096 & 0.0007 & 0.0538 & 0.0000 \\

MagFace FAR $10^{-3}$ ($<0.31$)        & 0.0618 & 0.0640 & 0.0522 & 0.0650 & 0.0634 & 0.0583 & 0.0705 & 0.0624 & 0.0633 & 0.0529 & 0.0652 & 0.0638 & 0.0579 & 0.0709 & 0.0587 \\
MagFace FAR $10^{-4}$ ($<0.18$)        & 0.0288 & 0.0276 & 0.0221 & 0.0287 & 0.0281 & 0.0267 & 0.0300 & 0.0284 & 0.0274 & 0.0223 & 0.0285 & 0.0280 & 0.0263 & 0.0299 & 0.0374 \\
MagFace FAR $10^{-5}$ ($<0.09$)        & 0.0099 & 0.0096 & 0.0076 & 0.0097 & 0.0099 & 0.0094 & 0.0102 & 0.0096 & 0.0096 & 0.0075 & 0.0100 & 0.0100 & 0.0090 & 0.0105 & 0.0159 \\

ElasticFace FAR $10^{-3}$ ($<0.80$)    & 0.0121 & 0.0754 & 0.0395 & 0.0785 & 0.0711 & 0.0124 & 0.1525 & 0.0121 & 0.0742 & 0.0394 & 0.0774 & 0.0726 & 0.0123 & 0.1519 & 0.0000 \\
ElasticFace FAR $10^{-4}$ ($<0.76$)    & 0.0057 & 0.0547 & 0.0256 & 0.0589 & 0.0469 & 0.0060 & 0.1253 & 0.0057 & 0.0554 & 0.0254 & 0.0586 & 0.0478 & 0.0057 & 0.1241 & 0.0000 \\
ElasticFace FAR $10^{-5}$ ($<0.70$)    & 0.0005 & 0.0158 & 0.0057 & 0.0185 & 0.0102 & 0.0006 & 0.0536 & 0.0003 & 0.0160 & 0.0052 & 0.0182 & 0.0094 & 0.0006 & 0.0528 & 0.0000 \\

CosFace FAR $10^{-3}$ ($<0.78$)        & 0.0107 & 0.0780 & 0.0419 & 0.0825 & 0.0635 & 0.0118 & 0.1552 & 0.0103 & 0.0757 & 0.0423 & 0.0814 & 0.0641 & 0.0118 & 0.1568 & 0.0000 \\
CosFace FAR $10^{-4}$ ($<0.72$)        & 0.0031 & 0.0431 & 0.0187 & 0.0464 & 0.0283 & 0.0033 & 0.1058 & 0.0028 & 0.0414 & 0.0188 & 0.0462 & 0.0275 & 0.0031 & 0.1066 & 0.0000 \\
CosFace FAR $10^{-5}$ ($<0.60$)        & 0.0000 & 0.0021 & 0.0005 & 0.0019 & 0.0006 & 0.0000 & 0.0108 & 0.0000 & 0.0019 & 0.0003 & 0.0020 & 0.0007 & 0.0000 & 0.0109 & 0.0000 \\

FaceNet FAR $10^{-3}$ ($<0.93$)        & 0.0194 & 0.0376 & 0.0281 & 0.0415 & 0.0400 & 0.0115 & 0.0704 & 0.0180 & 0.0371 & 0.0293 & 0.0403 & 0.0406 & 0.0108 & 0.0702 & 0.0000 \\
FaceNet FAR $10^{-4}$ ($<0.71$)        & 0.0004 & 0.0009 & 0.0006 & 0.0010 & 0.0008 & 0.0002 & 0.0025 & 0.0004 & 0.0008 & 0.0005 & 0.0010 & 0.0007 & 0.0002 & 0.0030 & 0.0000 \\
FaceNet FAR $10^{-5}$ ($<0.59$)        & 0.0000 & 0.0000 & 0.0000 & 0.0001 & 0.0000 & 0.0000 & 0.0002 & 0.0000 & 0.0000 & 0.0000 & 0.0001 & 0.0000 & 0.0000 & 0.0002 & 0.0000 \\
\cline{1-16}
Average FAR $10^{-3}$        & 0.0198 & 0.0636 & 0.0379 & \textbf{0.0681} & 0.0598 & 0.0188 & \emph{0.1201} & 0.0195 & 0.0623 & 0.0383 & \textbf{0.0674} & 0.0604 & 0.0184 & \emph{0.1211} & 0.0098 \\
Average FAR $10^{-4}$        & 0.0071 & 0.0326 & 0.0161 & \textbf{0.0359} & 0.0267 & 0.0071 & \emph{0.0753} & 0.0069 & 0.0322 & 0.0161 & \textbf{0.0358} & 0.0266 & 0.0069 & \emph{0.0759} & 0.0062 \\
Average FAR $10^{-5}$        & 0.0018 & 0.0079 & 0.0034 & \textbf{0.0089} & 0.0054 & 0.0018 & \emph{0.0253} & 0.0017 & 0.0078 & 0.0031 & \textbf{0.0090} & 0.0053 & 0.0018 & \emph{0.0252} & 0.0027 \\

\bottomrule

\end{tabular}
}
\end{center}
\end{table*}

\begin{table*}[htbp]
\caption{Quantitative experiments on FaceForensics++~\cite{faceforensics++} after tuning the victim system's identity encoder (ArcFace) on learned adversarial examples \cite{ReFace} and distortion attacks \cite{OpenAIRobust}. We evaluate successful identity retrieval (lower is better in terms of de-identification) after de-identifying the adversarial images using FIVA \cite{FIVA}. We show an average aggregation for each attack. Furthermore, we also show an average aggregation for each evaluation model per FAR threshold.}
\label{t:evaluation_id_tuned_distortion_system}
\begin{center}
\resizebox{1.0\textwidth}{!}{
\begin{tabular}{c|cccccc:c|cccccc:c|c}
\toprule
&\multicolumn{14}{c}{Gradient Calculation Models} \\
&\multicolumn{7}{c}{Cosine Distance Objective} &\multicolumn{7}{c}{Euclidean Distance Objective}\\
Evaluation Models            & ResNet & AdaFace & MagFace & ElasticFace & CosFace & FaceNet & ArcFace & ResNet & AdaFace & MagFace & ElasticFace & CosFace & FaceNet & ArcFace & No Attack \\
\cline{1-16}
ArcFace FAR $10^{-3}$ ($<0.76$)        & 0.0027 & 0.0567 & 0.0184 & 0.0639 & 0.0475 & 0.0040 & 0.1015 & 0.0027 & 0.0565 & 0.0189 & 0.0647 & 0.0467 & 0.0036 & 0.1027 & 0.0000 \\
ArcFace FAR $10^{-4}$ ($<0.69$)        & 0.0008 & 0.0301 & 0.0072 & 0.0350 & 0.0210 & 0.0013 & 0.0657 & 0.0006 & 0.0295 & 0.0072 & 0.0354 & 0.0208 & 0.0010 & 0.0668 & 0.0000 \\
ArcFace FAR $10^{-5}$ ($<0.62$)        & 0.0001 & 0.0064 & 0.0012 & 0.0081 & 0.0040 & 0.0001 & 0.0200 & 0.0000 & 0.0066 & 0.0012 & 0.0081 & 0.0039 & 0.0001 & 0.0215 & 0.0000 \\
\cdashline{1-16}
AdaFace FAR $10^{-3}$ ($<0.80$)        & 0.0075 & 0.0757 & 0.0303 & 0.0917 & 0.0736 & 0.0087 & 0.1382 & 0.0074 & 0.0747 & 0.0301 & 0.0918 & 0.0725 & 0.0085 & 0.1391 & 0.0000 \\
AdaFace FAR $10^{-4}$ ($<0.75$)        & 0.0032 & 0.0514 & 0.0166 & 0.0642 & 0.0444 & 0.0037 & 0.1012 & 0.0029 & 0.0515 & 0.0168 & 0.0639 & 0.0439 & 0.0037 & 0.1018 & 0.0000 \\
AdaFace FAR $10^{-5}$ ($<0.70$)        & 0.0005 & 0.0204 & 0.0044 & 0.0255 & 0.0145 & 0.0008 & 0.0470 & 0.0004 & 0.0197 & 0.0043 & 0.0260 & 0.0145 & 0.0007 & 0.0474 & 0.0000 \\

MagFace FAR $10^{-3}$ ($<0.31$)        & 0.0636 & 0.0656 & 0.0547 & 0.0680 & 0.0665 & 0.0604 & 0.0713 & 0.0649 & 0.0658 & 0.0541 & 0.0677 & 0.0673 & 0.0604 & 0.0708 & 0.0587 \\
MagFace FAR $10^{-4}$ ($<0.18$)        & 0.0295 & 0.0285 & 0.0231 & 0.0293 & 0.0290 & 0.0276 & 0.0305 & 0.0300 & 0.0279 & 0.0233 & 0.0295 & 0.0293 & 0.0271 & 0.0309 & 0.0374 \\
MagFace FAR $10^{-5}$ ($<0.09$)        & 0.0095 & 0.0093 & 0.0074 & 0.0097 & 0.0100 & 0.0090 & 0.0100 & 0.0095 & 0.0096 & 0.0072 & 0.0100 & 0.0096 & 0.0090 & 0.0104 & 0.0159 \\

ElasticFace FAR $10^{-3}$ ($<0.80$)    & 0.0091 & 0.0749 & 0.0304 & 0.0841 & 0.0700 & 0.0091 & 0.1328 & 0.0092 & 0.0758 & 0.0312 & 0.0856 & 0.0694 & 0.0094 & 0.1327 & 0.0000 \\
ElasticFace FAR $10^{-4}$ ($<0.76$)    & 0.0047 & 0.0577 & 0.0196 & 0.0672 & 0.0486 & 0.0047 & 0.1098 & 0.0048 & 0.0580 & 0.0201 & 0.0688 & 0.0486 & 0.0046 & 0.1100 & 0.0000 \\
ElasticFace FAR $10^{-5}$ ($<0.70$)    & 0.0005 & 0.0192 & 0.0043 & 0.0243 & 0.0141 & 0.0006 & 0.0476 & 0.0004 & 0.0189 & 0.0043 & 0.0248 & 0.0136 & 0.0006 & 0.0483 & 0.0000 \\

CosFace FAR $10^{-3}$ ($<0.78$)        & 0.0075 & 0.0803 & 0.0317 & 0.0882 & 0.0619 & 0.0088 & 0.1344 & 0.0073 & 0.0786 & 0.0312 & 0.0899 & 0.0615 & 0.0080 & 0.1376 & 0.0000 \\
CosFace FAR $10^{-4}$ ($<0.72$)        & 0.0031 & 0.0495 & 0.0156 & 0.0556 & 0.0329 & 0.0034 & 0.0942 & 0.0025 & 0.0481 & 0.0153 & 0.0573 & 0.0334 & 0.0032 & 0.0956 & 0.0000 \\
CosFace FAR $10^{-5}$ ($<0.60$)        & 0.0000 & 0.0028 & 0.0005 & 0.0040 & 0.0012 & 0.0000 & 0.0089 & 0.0000 & 0.0031 & 0.0003 & 0.0038 & 0.0012 & 0.0000 & 0.0093 & 0.0000 \\

FaceNet FAR $10^{-3}$ ($<0.93$)        & 0.0190 & 0.0398 & 0.0262 & 0.0434 & 0.0420 & 0.0116 & 0.0612 & 0.0174 & 0.0375 & 0.0265 & 0.0429 & 0.0422 & 0.0113 & 0.0609 & 0.0000 \\
FaceNet FAR $10^{-4}$ ($<0.71$)        & 0.0003 & 0.0009 & 0.0005 & 0.0011 & 0.0009 & 0.0002 & 0.0023 & 0.0003 & 0.0008 & 0.0004 & 0.0014 & 0.0010 & 0.0002 & 0.0024 & 0.0000 \\
FaceNet FAR $10^{-5}$ ($<0.59$)        & 0.0000 & 0.0000 & 0.0000 & 0.0001 & 0.0000 & 0.0000 & 0.0001 & 0.0000 & 0.0000 & 0.0000 & 0.0001 & 0.0000 & 0.0000 & 0.0001 & 0.0000 \\
\cline{1-16}
Average FAR $10^{-3}$        & 0.0182 & 0.0655 & 0.0320 & \textbf{0.0732} & 0.0603 & 0.0171 & \emph{0.1066} & 0.0182 & 0.0648 & 0.0320 & \textbf{0.0738} & 0.0599 & 0.0169 & \emph{0.1073} & 0.0098 \\
Average FAR $10^{-4}$        & 0.0069 & 0.0364 & 0.0138 & \textbf{0.0421} & 0.0295 & 0.0068 & \emph{0.0673} & 0.0069 & 0.0360 & 0.0139 & \textbf{0.0427} & 0.0295 & 0.0066 & \emph{0.0679} & 0.0062 \\
Average FAR $10^{-5}$        & 0.0018 & 0.0097 & 0.0030 & \textbf{0.0120} & 0.0073 & 0.0018 & \emph{0.0223} & 0.0017 & 0.0097 & 0.0029 & \textbf{0.0121} & 0.0071 & 0.0017 & \emph{0.0228} & 0.0027 \\

\bottomrule

\end{tabular}
}
\end{center}
\end{table*}


\begin{figure}[htbp]
\centering
\includegraphics[width=3.5in]{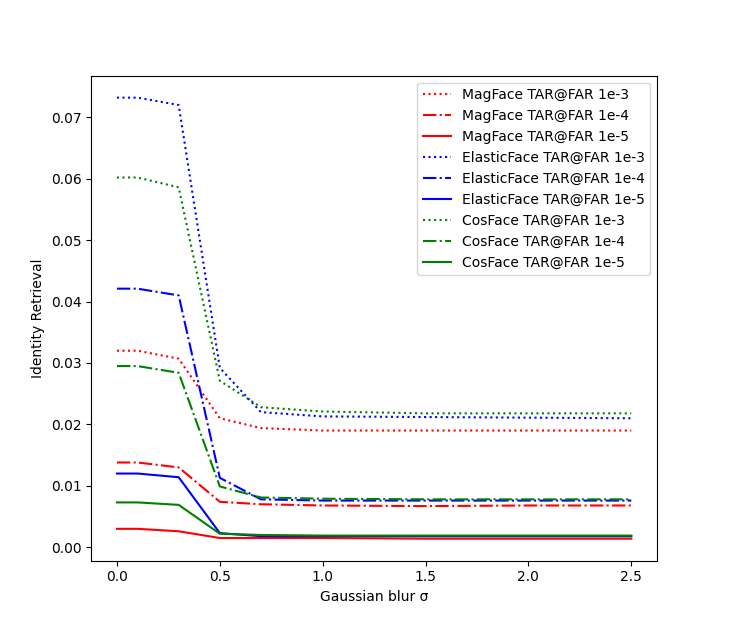}
\caption{
The figure shows how a low-pass filter (Gaussian blur) can filter away the adversarial noise and thus mitigate identity leakage. Figure shows the aggregated identity retrieval for three identity encoders as gradient calculation models: ElasticFace, MagFace and CosFace as we increase the $\sigma$ of the Gaussian blur kernel. $\sigma$ of 0 corresponds to the aggregated average values in Table \ref{t:evaluation_id_tuned_distortion_system}.}
\label{fig:low_pass_filter_effect}
\end{figure}

\begin{figure*}[htbp]
\centering
\includegraphics[width=\textwidth]{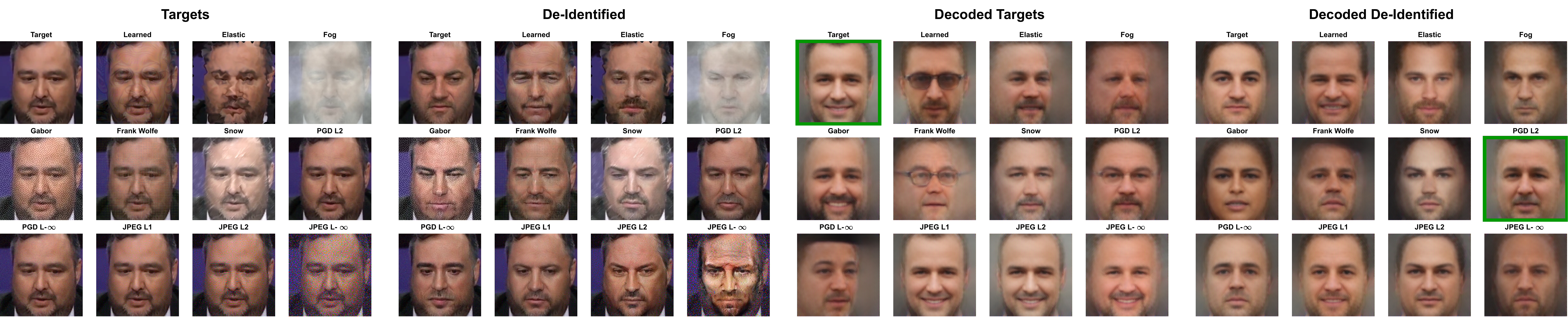}
\caption{
Right: Example of a target face and said target with the attacks applied. Left: Decoded embeddings of corresponding images using a DeepFaceDecoder \cite{DeepFaceDecoder}. The green boxes corresponds to an example of a match during identity retrieval.}
\label{fig:deepfacedecoding}
\end{figure*}

\begin{figure}[htbp]
\centering
\includegraphics[width=2.5in]{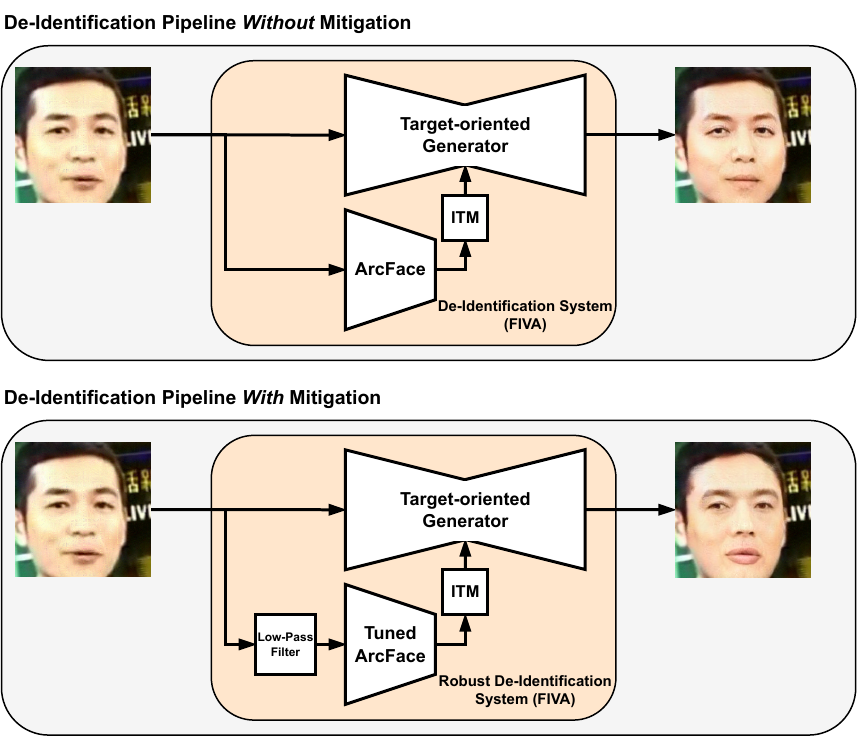}
\caption{
Illustration of the de-identification system without any mitigation against identity leakage due to adversarial attacks or distortions (See Table \ref{t:evaluation_id_baseline}) and the de-identification system with proposed mitigation defenses (See Table \ref{t:evaluation_id_tuned_distortion_system} and Figure \ref{fig:low_pass_filter_effect}). The low-pass filter corresponds to a Gaussian filter with $\sigma=1.5$ that acts as a de-noiser. ITM is the Identity Tracking Module used in the de-identification system, as introduced in \cite{FIVA}.}
\label{fig:mitigate_difference}
\end{figure}

\subsection{Quantitative Results}
\label{s:quant}
We provide comprehensive experimental results to provide a clear overview if transferability \cite{ADV_TRANS_1, ADV_TRANS_2} could lead to identity leakage and allow for a black-box-oriented type of attack against the victim system. The system we attack is an implementation of FIVA \cite{FIVA}. We follow their implementation, so the system contains an ArcFace identity encoder that is used to generate sampled fake identities from an Identity Tracking Module, which in turn is used to condition the generative model to produce a synthetic face. Due to our experimental space already being large (see Figure \ref{fig:attack_flow} and Table \ref{t:evaluation_id_baseline}), we limit which settings we test. We follow the original paper and set the margin for sampling the fake identity vector to $0.3$ in the Identity Tracking Module and turn off the tracking. We choose to turn off the tracking mechanism to simulate the scenario of the de-identification system encountering the face for the first time for each frame, providing more diverse synthetic faces. Table \ref{t:evaluation_id_baseline} contains quantitative results for successful identity retrieval with the six chosen identity encoders (CosFace, ArcFace, FaceNet, AdaFace, MagFace, ElasticFace) and the ResNet50 pretrained on ImageNet on the dataset FaceForensic++ \cite{faceforensics++}. This table compares different attacks for different configurations. Where a configuration would be a chosen identity encoder to calculate the gradients for the adversarial attack and one identity encoder that is used in the de-identification system. Each configuration is evaluated on all six identity encoders (even the one used for calculating gradient as a comparison) and for three acceptance thresholds for the FAR criteria of $10^{-3}$, $10^{-4}$ and $10^{-5}$. We calculate the threshold for each FAR value with randomly sampled genuine and imposter pairs from VGGFace2's train set \cite{vggface2}. The used thresholds are listed in Table \ref{t:evaluation_id_baseline}, Table \ref{t:evaluation_id_tuned_system} and Table \ref{t:evaluation_id_tuned_distortion_system}. The implementations in the advex-uar repository\footnote{https://github.com/ddkang/advex-uar} have an epsilon parameter that decides the strength of the adversarial noise. These are ambiguous, so we arbitrarily choose a strength that results in plausible images, where a non-desired image would make so much noise that you cannot really see that there is a face (see Table \ref{t:attack_strength} and Figure \ref{fig:attack_examples}).

\vspace{0.8mm}\noindent Referring to Table \ref{t:evaluation_id_baseline}, we can see that all the surrogate models are capable of causing identity leakage, with ElasticFace, CosFace and AdaFace having the strongest effect. This suggest that transferability is applicable here as well, meaning that the attack would not need access to the identity encoder of the victim system to cause identity leakage, especially if the surrogate model share similar traits (such as embedding to the same metric space). Naturally, if the attacker has access to the victim system identity encoder (ArcFace), the identity leakage is more effective. For the FaceNet model and a ResNet50 model pretrained on ImageNet as gradient calculation model, the effect on identity leakage is minor in this case, suggesting that models with similar metric space have a stronger effect, but not necessary. Table \ref{t:evaluation_id_baseline} also shows that using cosine distance or L2 distance for generating the adversarial examples as a similar effect. However, using a distance metric that matches the victim model's metric space will be slightly more effective.

\subsubsection{Robustness Tuning}
\textbf{Learned Attack Implementation Details:} The \emph{Learned} attack is an implementation from \cite{ReFace} with a few differences. The U-Net model is trained to produce a perturbation $p$ that is applied with $x_p = clamp(x_t + p * \epsilon_p)$. Where $x_t$ is the clean image, $x_p$ is the adversarial image, and $\epsilon_p$ is the strength of the perturbation. We set $\epsilon_p$ to 0.05 during training. We trained one version on ArcFace (see Figure \ref{fig:attack_flow}), where the identity loss $\mathcal{L}_{identity}$ (see Figure \ref{fig:attack_framwork}c) is maximizing the cosine distance between $x_t$ and $x_p$. Furthermore, we also apply a total variation loss $\mathcal{L}_{tv}$, with a weight of $\lambda_{tv} = 0.0001$. We use the AdamW \cite{AdamW} optimizer and set the learning rate to 0.0001, $\beta_0$ to 0.9, and $\beta_1$ to 0.999. Batch size is set to 32. We use VGGFace2 \cite{vggface2} for training.

\textbf{Finetuning Implementation Details:} All finetuned versions of the victim systems' identity encoder, ArcFace, are with $\lambda_{consistency} = 3$, $\lambda_{robustness} = 1$ and $\lambda_{union} = 2$. We use the AdamW \cite{AdamW} optimizer and set the learning rate to 0.0001, $\beta_0$ to 0.9, and $\beta_1$ to 0.999. Batch size is set to 32. For each adversarial example, we first apply the \emph{Learned} perturbation with a uniform sampled $\epsilon_{la}$ between 0.01 and 0.05 (see Figure \ref{fig:attack_framwork}b and \ref{fig:attack_framwork}c). Afterward we randomly apply one of the remaining attacks $N=3$ times, where there is a 50\% chance to apply it on the first iteration, 50\% on the second, and 30\% on the final (see Figure \ref{fig:attack_framwork}b). We uniformly sample a gain $\epsilon_{da}$ between 0.5 and 1.0, where 1.0 would result in strength listed in Table \ref{t:attack_strength}. During finetuning, we calculate the gradients for the attacks 10 times. We use VGGFace2 \cite{vggface2} for finetuning. This corresponds to the finetuned model evaluated in Table \ref{t:evaluation_id_tuned_distortion_system}. For the model evaluated in \ref{t:evaluation_id_tuned_system}, we only apply the \emph{Learned} attack.

Finetuning the de-identification systems' identity encoder does alleviate the identity leakage issues to an extent. Table \ref{t:evaluation_id_tuned_system} and Table \ref{t:evaluation_id_tuned_distortion_system} show the same results for when the victim system uses an identity encoder tuned on adversarial examples. Our experiments show that tuning the identity encoder in the victim de-identification system has some effect, where tuning on only \emph{Learned} \cite{ReFace} adversarial examples works better than combining \emph{Learned} with the remaining attacks (see Figure \ref{fig:attack_examples}).

To address the fact that there is still identity leakage present, we turn to simple signal processing. Considering the fact that the adversarial noise produced by these methods results in high frequency peaks in the image (see Figure \ref{fig:attack_examples}), we investigate if a low-pass filter in the form of a Gaussian filter in the de-identification pipeline that is applied before extracting the identity information (see Figure \ref{fig:mitigate_difference}) can act as a secondary defensive mechanism. The idea is that it will dampen the adversarial noise while the image is still interpretable for the identity encoder. Because ArcFace's embedding space is constrained to a hyper-sphere, the Gaussian blurring should have no effect on the actual de-identification model. This is because the de-identification model only uses the embedding for conditioning and \emph{not} the blurred image as input into the generator of FIVA (see Figure \ref{fig:low_pass_filter_effect}). As seen in Figure \ref{fig:low_pass_filter_effect}, we evaluate the identity leakage using the low-pass filter against attacks using MagFace, ElasticFace and CosFace for different $\sigma$ values of the Gaussian blur kernel. As theorized, the identity leakage suddenly drops. TAR at FAR $10^{-5}$ identity leakage drops to almost 0.

\subsubsection{Performance After tuning}
In Table \ref{t:evaluation_eer_LFW} we show the effect on performance on ArcFace the proposed robustness tuning has. The results suggest that the tuning has minimal effect. The tuned models do lose some performance for a FAR value of $10^{-4}$ and $10^{-5}$, but gain some for a FAR value of $10^{-3}$. Considering that this is in the context of face recognition and our goal is to make de-identification more robust to adversarial attacks, we consider this effect on performance negligible.

\begin{table*}[htbp]
\caption{Quantitative results and comparisons for the tuned models on LFW~\cite{LFW} 10-fold test protocol and FaceForensic++~\cite{faceforensics++}. We show the equal error rate (EER), false rejection rate (FRR) for a FAR of $10^{-3}$, $10^{-4}$ and $10^{-5}$, and the accuracy for a FAR of $10^{-3}$, $10^{-4}$ and $10^{-5}$.}
\label{t:evaluation_eer_LFW}
\begin{center}
\begin{tabular}{cc|ccccccc}
\toprule
&Model             & EER & FRR \tiny (FAR $10^{-3}$) & FRR \tiny (FAR $10^{-4}$) & FRR \tiny (FAR $10^{-5}$) & Acc \tiny (FAR $10^{-3}$) & Acc \tiny (FAR $10^{-4}$) & Acc \tiny (FAR $10^{-5}$) \\
\cline{2-9}
&CosFace                                         & 4.27\% & 4.97\% & 5.07\% & 5.07\% & 97.47\% & 97.47\% & 97.47\% \\
&FaceNet                                         & 4.37\% & 7.80\% & 38.43\% & 38.43\% & 96.05\% & 80.78\% & 80.78\% \\
\parbox[t]{2mm}{\multirow{1}{*}{\rotatebox[origin=c]{90}{\scriptsize LFW}}}
&ArcFace                                         & 4.44\% & 5.00\% & 5.03\% & 5.03\% & 97.45\% & 97.48\% & 97.48\% \\
\cdashline{2-9}
&ArcFace \tiny(Learned attacks)                  & 0.30\% & 4.87\% & 6.67\% & 6.67\% & 97.53\% & 96.67\% & 96.67\% \\
&ArcFace \tiny(Learned + distortion attacks)     & 0.30\% & 4.90\% & 7.00\% & 7.00\% & 97.52\% & 96.50\% & 96.50\% \\
\cline{2-9}
\parbox[t]{2mm}{\multirow{5}{*}{\rotatebox[origin=c]{90}{\scriptsize FaceForensic++}}}
&CosFace                                         & 0.42\% & 0.52\% & 0.65\% & 11.65\% & 99.69\% & 97.47\% & 94.18\% \\
&FaceNet                                         & 0.69\% & 1.20\% & 3.01\% & 12.94\% & 99.35\% & 98.49\% & 93.53\% \\
&ArcFace                                         & 0.47\% & 0.57\% & 0.75\% & 10.94\% & 99.67\% & 99.62\% & 94.53\% \\
\cdashline{2-9}
&ArcFace \tiny(Learned attacks)                  & 0.47\% & 0.59\% & 0.81\% & 16.73\% & 99.66\% & 99.59\% & 91.64\% \\
&ArcFace \tiny(Learned + distortion attacks)     & 0.49\% & 0.53\% & 0.76\% & 16.60\% & 99.69\% & 99.62\% & 91.70\% \\
\bottomrule
\end{tabular}

\end{center}
\end{table*}

\subsection{Qualitative Results}
\label{s:qual}

For qualitative results, we utilize DeepFaceDecoder \cite{DeepFaceDecoder} to decode ArcFace embeddings. We choose to only visualize ArcFace embeddings because we only need to use one identity encoder's embeddings to decode after any attacks have been applied. To clarify further, no matter which encoder was used to produce the attack, it is still enough to visualize the effect with ArcFace and its corresponding DeepFaceDecoder. We use this tool as a human-friendly visualization tool for investigating when the identity leaks. As shown in Figure \ref{fig:match_non_match} we show examples from FaceForensics++ \cite{faceforensics++} of when three different attacks (\emph{Frank Wolfe}, \emph{PGD L2} and \emph{PGD L-}$\infty$) cause a match using CosFace and when it does not. While the identity leakage is not as noticeable in the "de-identified" image, we can better observe the effect by looking into the decoded corresponding ArcFace embedding. The same goes for applying the low-pass de-noising filter, which prevents the attacks from leaking the identity in the present examples. Figure \ref{fig:deepfacedecoding} further illustrates how the decoded face changes due to the applied attacks, due to de-identification and the combination.

Furthermore, Figure \ref{fig:deepfacedecoding} also illustrates the visual effect the attacks have on the resulting de-identified version. To clarify, we can see that most attacks still cause the de-identification system to produce a convincing face except for \emph{JPEG L-}$\infty$, which does degrade the realism of the synthetic face significantly. This is arguable not of significant concern, as we have noted that \emph{PGD L-}$\infty$ does not cause significant identity leakage (see supplementary materials).


\section{Conclusion}
In this work, we investigated identity leakage in realistic de-identification caused by adversarial noise targeted at the system's identity encoder. FIVA \cite{FIVA} was used as the reference victim system as the measurement tool for the effect on the task of de-identification. This means that de-identification methods that leverage an identity encoder would suffer the same vulnerabilities. We demonstrated that an adversary could generate adversarial examples with a surrogate model (e.g., CosFace) to cause ArcFace to extract identity  in such a way that the resulting de-identified image leakages the identity. Thus causing a potential biometric system to produce a true acceptance of the original identity. For example, the aggregated results using AdaFace as a surrogate model show a match in a biometric verification system of 2.45\% of the time using ElasticFace with a FAR criterion of $10^{-5}$. We show that the adversary can achieve identity leakage even if the surrogate model is not similar to the victim model. As using ResNet trained on ImageNet as a surrogate model can cause a match of 2.98\% in a facial recognition system with MagFace with a FAR criterion of $10^{-4}$. Furthermore, we demonstrate a couple of defense mechanisms to deal with adversarial noise, whether it is caused by a bad actor or real world distortions. First, a low-pass filter, specifically a Gaussian filter, can dampen high frequency adversarial noise to nullify the identity leakage almost completely for a biometric system with high security and certainty demand (FAR criteria of $10^{-5}$ or lower). Finally, we address the remaining distortion-based adversarial attacks by distilling the de-identification system's identity encoder on adversarial examples. We believe this will push the practicality of de-identification further. Our empirical studies show that transferability is possible even if the model's task does not match with the victim model for a wide range of attack scenarios. This demonstrates that adversarial attacks could be used to both probe and attack a de-identification system in a black-box manner.

For future directions, there are several paths of interest. Firstly, there are numerous more different types of adversarial attacks that could be investigated, particularly the real-world patch-based ones \cite{AdvHat, AdvSticker}. Secondly, looking into iterative robustness, where we would test the system's robustness against an adversary that adapts to the mitigation proposed in this paper.

\appendices
\section{Non-Aggregated Results}
In the main manuscript, we provided aggregated quantitative results of the success rate of identity retrieval for each attack. These results were aggregated by using different surrogate models to calculate the gradients for each attack. In this supplementary material, we show the results for each surrogate model, for each attack, without any aggregation. Shown in Table \ref{t:resnet_details}, \ref{t:adaface_details}, \ref{t:magface_details}, \ref{t:elasticface_details}, \ref{t:cosface_details}, \ref{t:facenet_details}, \ref{t:arcface_details}, \ref{t:tuned_resnet_details}, \ref{t:tuend_adaface_details}, \ref{t:tuend_magface_details}, \ref{t:tuend_elastiface_details}, \ref{t:tuend_cosface_details}, \ref{t:tuend_facenet_details}, \ref{t:tuend_arcface_details}, \ref{t:dis_tuned_resnet_details}, \ref{t:dis_tuned_adaface_details}, \ref{t:dis_tuned_magface_details}, \ref{t:dis_tuned_elasticface_details}, \ref{t:dis_tuned_cosface_details}, \ref{t:dis_tuned_facenet_details}, \ref{t:dis_tuned_arcface_details}, which includes results for attacking the system with tuned identity encoders.
\begin{table*}[h!]
\begin{center}
\resizebox{1.0\textwidth}{!}{
\begin{tabular}{c|cccccccccc|cccccccccc}
\toprule
&\multicolumn{20}{c}{Attack Type} \\
&\multicolumn{10}{c}{Cosine Distance Objective} &\multicolumn{10}{c}{Euclidean Distance Objective}\\
Evaluation Models             & Snow & Elastic & Fog & Frank Wolfe & Gabor & JPEG L1 & JPEG L2 & JPEG L$\infty$ & PGD L2 & PGD L$\infty$ & Snow & Elastic & Fog & Frank Wolfe & Gabor & JPEG L1 & JPEG L2 & JPEG L$\infty$ & PGD L2 & PGD L$\infty$\\
\cline{1-21}
ArcFace FAR $10^{-3}$ ($<0.76$)        & 0.0021 & 0.0054 & 0.0028 & 0.0024 & 0.002 & 0.0019 & 0.0007 & 0.0035 & 0.0043 & 0.0014 & 0.0013 & 0.0042 & 0.0052 & 0.0029 & 0.0009 & 0.0019 & 0.0003 & 0.0031 & 0.0034 & 0.0022 \\
ArcFace FAR $10^{-4}$ ($<0.69$)        & 0.0002 & 0.0022 & 0.0007 & 0.0005 & 0.0004 & 0.0004 & 0.0001 & 0.0005 & 0.0009 & 0.0001 & 0.0001 & 0.0007 & 0.0015 & 0.001 & 0.0002 & 0.0004 & 0.0000 & 0.0002 & 0.0011 & 0.0003 \\
ArcFace FAR $10^{-5}$ ($<0.62$)        & 0.0000 & 0.0003 & 0.0000 & 0.0001 & 0.0000 & 0.0000 & 0.0000 & 0.0000 & 0.0002 & 0.0000 & 0.0000 & 0.0000 & 0.0000 & 0.0000 & 0.0000 & 0.0000 & 0.0000 & 0.0001 & 0.0000 & 0.0001 \\
\cdashline{1-21}
AdaFace FAR $10^{-3}$ ($<0.80$)        & 0.0029 & 0.0161 & 0.0087 & 0.0082 & 0.0036 & 0.0086 & 0.0033 & 0.004 & 0.0144 & 0.0078 & 0.0047 & 0.0167 & 0.0087 & 0.0089 & 0.0025 & 0.0071 & 0.0016 & 0.0035 & 0.0107 & 0.0099 \\
AdaFace FAR $10^{-4}$ ($<0.75$)        & 0.0012 & 0.0091 & 0.0032 & 0.0033 & 0.0011 & 0.0033 & 0.002 & 0.0005 & 0.0072 & 0.0034 & 0.0011 & 0.0082 & 0.0034 & 0.0034 & 0.0007 & 0.002 & 0.0003 & 0.0004 & 0.0055 & 0.0037 \\
\hline

\end{tabular}
}
\end{center}
\caption{Quantitative experiments on FaceForensics++~\cite{faceforensics++}. We show the detailed results for each attack using ResNet50 trained on ImageNet as the surrogate model.}
\label{t:resnet_details}
\end{table*}

\begin{table*}[h!]
\begin{center}
\resizebox{1.0\textwidth}{!}{
\begin{tabular}{c|cccccccccc|cccccccccc}
\toprule
&\multicolumn{20}{c}{Attack Type} \\
&\multicolumn{10}{c}{Cosine Distance Objective} &\multicolumn{10}{c}{Euclidean Distance Objective}\\
Evaluation Models             & Snow & Elastic & Fog & Frank Wolfe & Gabor & JPEG L1 & JPEG L2 & JPEG L$\infty$ & PGD L2 & PGD L$\infty$ & Snow & Elastic & Fog & Frank Wolfe & Gabor & JPEG L1 & JPEG L2 & JPEG L$\infty$ & PGD L2 & PGD L$\infty$\\
\cline{1-21}
ArcFace FAR $10^{-3}$ ($<0.76$)     & 0.0063 & 0.0391 & 0.0146 & 0.3156 & 0.044 & 0.0018 & 0.0014 & 0.0008 & 0.004 & 0.2191 & 0.0076 & 0.0378 & 0.0165 & 0.3098 & 0.0424 & 0.0022 & 0.0003 & 0.0009 & 0.0046 & 0.218 \\
ArcFace FAR $10^{-4}$ ($<0.69$)     & 0.0023 & 0.0107 & 0.0046 & 0.1839 & 0.0157 & 0.0003 & 0.0003 & 0.0001 & 0.0011 & 0.1239 & 0.0021 & 0.0111 & 0.0042 & 0.1753 & 0.0149 & 0.0002 & 0.0002 & 0.0003 & 0.0009 & 0.12 \\
ArcFace FAR $10^{-5}$ ($<0.62$)     & 0.0000 & 0.0006 & 0.0006 & 0.0428 & 0.0025 & 0.0000 & 0.0001 & 0.0000 & 0.0000 & 0.0287 & 0.0000 & 0.001 & 0.0002 & 0.0359 & 0.0024 & 0.0000 & 0.0000 & 0.0000 & 0.0001 & 0.026 \\
\cdashline{1-21}
AdaFace FAR $10^{-3}$ ($<0.80$)     & 0.0044 & 0.0323 & 0.0084 & 0.4114 & 0.0777 & 0.0085 & 0.0022 & 0.001 & 0.0127 & 0.3352 & 0.0045 & 0.0307 & 0.009 & 0.4142 & 0.0722 & 0.0072 & 0.0014 & 0.0012 & 0.0118 & 0.328 \\
AdaFace FAR $10^{-4}$ ($<0.75$)     & 0.0016 & 0.0148 & 0.0031 & 0.3003 & 0.0377 & 0.0036 & 0.0008 & 0.0003 & 0.0062 & 0.2405 & 0.0011 & 0.0154 & 0.0028 & 0.3005 & 0.0358 & 0.0027 & 0.0004 & 0.0000 & 0.0059 & 0.2355 \\
\hline

\end{tabular}
}
\end{center}
\caption{Quantitative experiments on FaceForensics++~\cite{faceforensics++}. We show the detailed results for each attack using AdaFace as the surrogate model.}
\label{t:adaface_details}
\end{table*}

\begin{table*}[h!]
\begin{center}
\resizebox{1.0\textwidth}{!}{
\begin{tabular}{c|cccccccccc|cccccccccc}
\toprule
&\multicolumn{20}{c}{Attack Type} \\
&\multicolumn{10}{c}{Cosine Distance Objective} &\multicolumn{10}{c}{Euclidean Distance Objective}\\
Evaluation Models             & Snow & Elastic & Fog & Frank Wolfe & Gabor & JPEG L1 & JPEG L2 & JPEG L$\infty$ & PGD L2 & PGD L$\infty$ & Snow & Elastic & Fog & Frank Wolfe & Gabor & JPEG L1 & JPEG L2 & JPEG L$\infty$ & PGD L2 & PGD L$\infty$\\
\cline{1-21}
ArcFace FAR $10^{-3}$ ($<0.76$)     & 0.0065 & 0.0289 & 0.019 & 0.1897 & 0.009 & 0.0031 & 0.001 & 0.0125 & 0.004 & 0.031 & 0.0063 & 0.0338 & 0.0211 & 0.183 & 0.0103 & 0.0024 & 0.0007 & 0.0127 & 0.0033 & 0.0231 \\
ArcFace FAR $10^{-4}$ ($<0.69$)     & 0.0019 & 0.0095 & 0.0053 & 0.0901 & 0.0032 & 0.0009 & 0.0005 & 0.0052 & 0.001 & 0.0121 & 0.0024 & 0.0126 & 0.0073 & 0.0821 & 0.0019 & 0.0007 & 0.0002 & 0.0047 & 0.001 & 0.0072 \\
ArcFace FAR $10^{-5}$ ($<0.62$)     & 0.0002 & 0.0009 & 0.0004 & 0.0158 & 0.0003 & 0.0000 & 0.0001 & 0.0003 & 0.0002 & 0.0016 & 0.0000 & 0.0014 & 0.0008 & 0.0131 & 0.0002 & 0.0001 & 0.0000 & 0.0005 & 0.0001 & 0.0007 \\
\cdashline{1-21}
AdaFace FAR $10^{-3}$ ($<0.80$)     & 0.0109 & 0.0565 & 0.0272 & 0.2361 & 0.0182 & 0.0097 & 0.0021 & 0.0142 & 0.0132 & 0.0623 & 0.0098 & 0.0578 & 0.028 & 0.2348 & 0.0184 & 0.0094 & 0.001 & 0.0149 & 0.0112 & 0.0573 \\
AdaFace FAR $10^{-4}$ ($<0.75$)     & 0.0043 & 0.0323 & 0.0126 & 0.154 & 0.0072 & 0.0039 & 0.0007 & 0.0042 & 0.0065 & 0.0363 & 0.0039 & 0.0336 & 0.0131 & 0.1471 & 0.0077 & 0.0043 & 0.0003 & 0.0042 & 0.004 & 0.032 \\
\hline

\end{tabular}
}
\end{center}
\caption{Quantitative experiments on FaceForensics++~\cite{faceforensics++}. We show the detailed results for each attack using MagFace as the surrogate model.}
\label{t:magface_details}
\end{table*}

\begin{table*}[h!]
\begin{center}
\resizebox{1.0\textwidth}{!}{
\begin{tabular}{c|cccccccccc|cccccccccc}
\toprule
&\multicolumn{20}{c}{Attack Type} \\
&\multicolumn{10}{c}{Cosine Distance Objective} &\multicolumn{10}{c}{Euclidean Distance Objective}\\
Evaluation Models             & Snow & Elastic & Fog & Frank Wolfe & Gabor & JPEG L1 & JPEG L2 & JPEG L$\infty$ & PGD L2 & PGD L$\infty$ & Snow & Elastic & Fog & Frank Wolfe & Gabor & JPEG L1 & JPEG L2 & JPEG L$\infty$ & PGD L2 & PGD L$\infty$\\
\cline{1-21}
ArcFace FAR $10^{-3}$ ($<0.76$)     & 0.0138 & 0.0439 & 0.0174 & 0.3667 & 0.0726 & 0.004 & 0.0008 & 0.0004 & 0.0029 & 0.3062 & 0.013 & 0.0439 & 0.0181 & 0.3563 & 0.0779 & 0.0044 & 0.0004 & 0.0004 & 0.0036 & 0.3106 \\
ArcFace FAR $10^{-4}$ ($<0.69$)     & 0.0044 & 0.0153 & 0.0044 & 0.2262 & 0.0336 & 0.0012 & 0.0004 & 0.0000 & 0.0005 & 0.1953 & 0.0053 & 0.016 & 0.0051 & 0.222 & 0.031 & 0.0011 & 0.0001 & 0.0001 & 0.0011 & 0.1899 \\
ArcFace FAR $10^{-5}$ ($<0.62$)     & 0.0003 & 0.0004 & 0.0002 & 0.0596 & 0.0049 & 0.0000 & 0.0000 & 0.0000 & 0.0001 & 0.056 & 0.0005 & 0.0012 & 0.0005 & 0.056 & 0.0038 & 0.0000 & 0.0000 & 0.0000 & 0.0002 & 0.0508 \\
\cdashline{1-21}
AdaFace FAR $10^{-3}$ ($<0.80$)     & 0.0132 & 0.0858 & 0.0197 & 0.4566 & 0.1153 & 0.0104 & 0.0022 & 0.0011 & 0.0124 & 0.4297 & 0.0132 & 0.0888 & 0.0194 & 0.4412 & 0.1113 & 0.0116 & 0.0016 & 0.0007 & 0.0127 & 0.4182 \\
AdaFace FAR $10^{-4}$ ($<0.75$)     & 0.0062 & 0.0469 & 0.0071 & 0.3527 & 0.0586 & 0.0063 & 0.001 & 0.0000 & 0.0046 & 0.338 & 0.0069 & 0.0484 & 0.0088 & 0.3465 & 0.0591 & 0.007 & 0.0005 & 0.0002 & 0.0059 & 0.3239 \\
\hline

\end{tabular}
}
\end{center}
\caption{Quantitative experiments on FaceForensics++~\cite{faceforensics++}. We show the detailed results for each attack using ElasticFace as the surrogate model.}
\label{t:elasticface_details}
\end{table*}

\begin{table*}[h!]
\begin{center}
\resizebox{1.0\textwidth}{!}{
\begin{tabular}{c|cccccccccc|cccccccccc}
\toprule
&\multicolumn{20}{c}{Attack Type} \\
&\multicolumn{10}{c}{Cosine Distance Objective} &\multicolumn{10}{c}{Euclidean Distance Objective}\\
Evaluation Models             & Snow & Elastic & Fog & Frank Wolfe & Gabor & JPEG L1 & JPEG L2 & JPEG L$\infty$ & PGD L2 & PGD L$\infty$ & Snow & Elastic & Fog & Frank Wolfe & Gabor & JPEG L1 & JPEG L2 & JPEG L$\infty$ & PGD L2 & PGD L$\infty$\\
\cline{1-21}
ArcFace FAR $10^{-3}$ ($<0.76$)     & 0.0163 & 0.0541 & 0.0154 & 0.211 & 0.1082 & 0.0049 & 0.0007 & 0.0004 & 0.0037 & 0.3654 & 0.0182 & 0.0497 & 0.0179 & 0.2131 & 0.1008 & 0.0053 & 0.0003 & 0.0005 & 0.0038 & 0.3615 \\
ArcFace FAR $10^{-4}$ ($<0.69$)     & 0.0045 & 0.0164 & 0.0041 & 0.1142 & 0.0452 & 0.0004 & 0.0005 & 0.0001 & 0.001 & 0.2372 & 0.0035 & 0.014 & 0.0061 & 0.1123 & 0.0476 & 0.0016 & 0.0000 & 0.0001 & 0.0011 & 0.2378 \\
ArcFace FAR $10^{-5}$ ($<0.62$)     & 0.0008 & 0.0007 & 0.0003 & 0.0275 & 0.008 & 0.0000 & 0.0000 & 0.0000 & 0.0003 & 0.0665 & 0.0003 & 0.001 & 0.0003 & 0.0263 & 0.0094 & 0.0000 & 0.0000 & 0.0000 & 0.0002 & 0.0671 \\
\cdashline{1-21}
AdaFace FAR $10^{-3}$ ($<0.80$)     & 0.0167 & 0.1053 & 0.0187 & 0.2703 & 0.157 & 0.0126 & 0.0019 & 0.0002 & 0.0116 & 0.4999 & 0.0152 & 0.1009 & 0.022 & 0.2713 & 0.1454 & 0.0129 & 0.0024 & 0.0007 & 0.0127 & 0.489 \\
AdaFace FAR $10^{-4}$ ($<0.75$)     & 0.0062 & 0.0537 & 0.0075 & 0.1859 & 0.09 & 0.0066 & 0.0006 & 0.0000 & 0.0056 & 0.3946 & 0.0048 & 0.049 & 0.0094 & 0.1879 & 0.0831 & 0.0072 & 0.001 & 0.0002 & 0.0062 & 0.3861 \\
\hline

\end{tabular}
}
\end{center}
\caption{Quantitative experiments on FaceForensics++~\cite{faceforensics++}. We show the detailed results for each attack using CosFace as the surrogate model.}
\label{t:cosface_details}
\end{table*}

\begin{table*}[h!]
\begin{center}
\resizebox{1.0\textwidth}{!}{
\begin{tabular}{c|cccccccccc|cccccccccc}
\toprule
&\multicolumn{20}{c}{Attack Type} \\
&\multicolumn{10}{c}{Cosine Distance Objective} &\multicolumn{10}{c}{Euclidean Distance Objective}\\
Evaluation Models             & Snow & Elastic & Fog & Frank Wolfe & Gabor & JPEG L1 & JPEG L2 & JPEG L$\infty$ & PGD L2 & PGD L$\infty$ & Snow & Elastic & Fog & Frank Wolfe & Gabor & JPEG L1 & JPEG L2 & JPEG L$\infty$ & PGD L2 & PGD L$\infty$\\
\cline{1-21}
ArcFace FAR $10^{-3}$ ($<0.76$)     & 0.0012 & 0.0193 & 0.0042 & 0.0052 & 0.0007 & 0.0022 & 0.0005 & 0.0079 & 0.003 & 0.0015 & 0.0018 & 0.0157 & 0.0045 & 0.0038 & 0.0006 & 0.0036 & 0.0006 & 0.0061 & 0.0037 & 0.002 \\
ArcFace FAR $10^{-4}$ ($<0.69$)     & 0.0005 & 0.007 & 0.0006 & 0.0019 & 0.0001 & 0.0005 & 0.0003 & 0.0016 & 0.0009 & 0.0003 & 0.0007 & 0.0046 & 0.0008 & 0.0004 & 0.0001 & 0.0009 & 0.0003 & 0.0009 & 0.0012 & 0.0006 \\
ArcFace FAR $10^{-5}$ ($<0.62$)     & 0.0001 & 0.0006 & 0.0000 & 0.0002 & 0.0000 & 0.0000 & 0.0000 & 0.0001 & 0.0002 & 0.0000 & 0.0001 & 0.0004 & 0.0000 & 0.0000 & 0.0000 & 0.0000 & 0.0001 & 0.0000 & 0.0001 & 0.0 \\
\cdashline{1-21}
AdaFace FAR $10^{-3}$ ($<0.80$)     & 0.0013 & 0.0308 & 0.0057 & 0.0138 & 0.0026 & 0.0107 & 0.0037 & 0.0065 & 0.0127 & 0.0059 & 0.0029 & 0.032 & 0.0047 & 0.0092 & 0.0011 & 0.0109 & 0.0016 & 0.0074 & 0.0125 & 0.0068 \\
AdaFace FAR $10^{-4}$ ($<0.75$)     & 0.0004 & 0.0156 & 0.002 & 0.0058 & 0.0006 & 0.0047 & 0.0014 & 0.0015 & 0.0047 & 0.0029 & 0.0006 & 0.0177 & 0.0015 & 0.0029 & 0.0003 & 0.0057 & 0.0011 & 0.0022 & 0.0054 & 0.0026 \\
\hline

\end{tabular}
}
\end{center}
\caption{Quantitative experiments on FaceForensics++~\cite{faceforensics++}. We show the detailed results for each attack using FaceNet as the surrogate model.}
\label{t:facenet_details}
\end{table*}

\begin{table*}[h!]
\begin{center}
\resizebox{1.0\textwidth}{!}{
\begin{tabular}{c|cccccccccc|cccccccccc}
\toprule
&\multicolumn{20}{c}{Attack Type} \\
&\multicolumn{10}{c}{Cosine Distance Objective} &\multicolumn{10}{c}{Euclidean Distance Objective}\\
Evaluation Models             & Snow & Elastic & Fog & Frank Wolfe & Gabor & JPEG L1 & JPEG L2 & JPEG L$\infty$ & PGD L2 & PGD L$\infty$ & Snow & Elastic & Fog & Frank Wolfe & Gabor & JPEG L1 & JPEG L2 & JPEG L$\infty$ & PGD L2 & PGD L$\infty$\\
\cline{1-21}
ArcFace FAR $10^{-3}$ ($<0.76$)     & 0.0341 & 0.1011 & 0.0398 & 0.8273 & 0.3305 & 0.0033 & 0.0004 & 0.0009 & 0.0031 & 0.8464 & 0.0349 & 0.0991 & 0.0387 & 0.8182 & 0.3266 & 0.0023 & 0.0005 & 0.0006 & 0.0038 & 0.8442 \\
ArcFace FAR $10^{-4}$ ($<0.69$)     & 0.0111 & 0.0295 & 0.011 & 0.7212 & 0.2066 & 0.0011 & 0.0000 & 0.0003 & 0.0006 & 0.7568 & 0.0139 & 0.0299 & 0.0098 & 0.7131 & 0.201 & 0.0004 & 0.0000 & 0.0002 & 0.0012 & 0.7558 \\
ArcFace FAR $10^{-5}$ ($<0.62$)     & 0.001 & 0.0025 & 0.0012 & 0.374 & 0.0622 & 0.0000 & 0.0000 & 0.0000 & 0.0000 & 0.4247 & 0.0018 & 0.0029 & 0.0015 & 0.376 & 0.0584 & 0.0000 & 0.0000 & 0.0000 & 0.0001 & 0.4125 \\
\cdashline{1-21}
AdaFace FAR $10^{-3}$ ($<0.80$)     & 0.071 & 0.3416 & 0.0869 & 0.8474 & 0.4127 & 0.0094 & 0.0023 & 0.0009 & 0.0119 & 0.884 & 0.0727 & 0.3488 & 0.0873 & 0.8374 & 0.4128 & 0.0074 & 0.002 & 0.0007 & 0.013 & 0.8871 \\
AdaFace FAR $10^{-4}$ ($<0.75$)     & 0.0353 & 0.2285 & 0.0444 & 0.7856 & 0.3016 & 0.0049 & 0.0007 & 0.0004 & 0.0052 & 0.8439 & 0.0352 & 0.2332 & 0.0411 & 0.7759 & 0.2984 & 0.0044 & 0.0007 & 0.0001 & 0.0057 & 0.8449 \\
\hline

\end{tabular}
}
\end{center}
\caption{Quantitative experiments on FaceForensics++~\cite{faceforensics++}. We show the detailed results for each attack using ArcFace as the surrogate model.}
\label{t:arcface_details}
\end{table*}

\begin{table*}[h!]
\begin{center}
\resizebox{1.0\textwidth}{!}{
\begin{tabular}{c|cccccccccc|cccccccccc}
\toprule
&\multicolumn{20}{c}{Attack Type} \\
&\multicolumn{10}{c}{Cosine Distance Objective} &\multicolumn{10}{c}{Euclidean Distance Objective}\\
Evaluation Models             & Snow & Elastic & Fog & Frank Wolfe & Gabor & JPEG L1 & JPEG L2 & JPEG L$\infty$ & PGD L2 & PGD L$\infty$ & Snow & Elastic & Fog & Frank Wolfe & Gabor & JPEG L1 & JPEG L2 & JPEG L$\infty$ & PGD L2 & PGD L$\infty$\\
\cline{1-21}
ArcFace FAR $10^{-3}$ ($<0.76$)     & 0.0021 & 0.0044 & 0.0038 & 0.0015 & 0.001 & 0.0011 & 0.0005 & 0.0233 & 0.0037 & 0.0014 & 0.002 & 0.0035 & 0.0035 & 0.002 & 0.0008 & 0.0018 & 0.0006 & 0.0211 & 0.0035 & 0.0016 \\
ArcFace FAR $10^{-4}$ ($<0.69$)     & 0.0004 & 0.0022 & 0.0011 & 0.0003 & 0.0001 & 0.0002 & 0.0001 & 0.0029 & 0.0002 & 0.0000 & 0.0000 & 0.0004 & 0.0007 & 0.0005 & 0.0001 & 0.0001 & 0.0000 & 0.0033 & 0.0001 & 0.0001 \\
ArcFace FAR $10^{-5}$ ($<0.62$)     & 0.0000 & 0.0002 & 0.0001 & 0.0001 & 0.0001 & 0.0000 & 0.0000 & 0.0000 & 0.0001 & 0.0000 & 0.0000 & 0.0000 & 0.0000 & 0.0000 & 0.0000 & 0.0000 & 0.0000 & 0.0001 & 0.0000 & 0.0001 \\
\cdashline{1-21}
AdaFace FAR $10^{-3}$ ($<0.80$)     & 0.0059 & 0.0165 & 0.0104 & 0.0099 & 0.0029 & 0.0079 & 0.0034 & 0.0246 & 0.0148 & 0.0091 & 0.0055 & 0.0148 & 0.0089 & 0.0098 & 0.0034 & 0.0078 & 0.0056 & 0.023 & 0.0138 & 0.0088 \\
AdaFace FAR $10^{-4}$ ($<0.75$)     & 0.0027 & 0.0082 & 0.0042 & 0.0028 & 0.0009 & 0.0032 & 0.0009 & 0.0046 & 0.0071 & 0.0034 & 0.0009 & 0.0069 & 0.0036 & 0.0041 & 0.001 & 0.0028 & 0.0023 & 0.0045 & 0.0064 & 0.0037 \\
\hline

\end{tabular}
}
\end{center}
\caption{Quantitative experiments on FaceForensics++~\cite{faceforensics++} after tuning the victim system's identity encoder (ArcFace) on learned adversarial examples \cite{ReFace}. We show the detailed results for each attack using ResNet50 trained on ImageNet as the surrogate model.}
\label{t:tuned_resnet_details}
\end{table*}

\begin{table*}[h!]
\begin{center}
\resizebox{1.0\textwidth}{!}{
\begin{tabular}{c|cccccccccc|cccccccccc}
\toprule
&\multicolumn{20}{c}{Attack Type} \\
&\multicolumn{10}{c}{Cosine Distance Objective} &\multicolumn{10}{c}{Euclidean Distance Objective}\\
Evaluation Models             & Snow & Elastic & Fog & Frank Wolfe & Gabor & JPEG L1 & JPEG L2 & JPEG L$\infty$ & PGD L2 & PGD L$\infty$ & Snow & Elastic & Fog & Frank Wolfe & Gabor & JPEG L1 & JPEG L2 & JPEG L$\infty$ & PGD L2 & PGD L$\infty$\\
\cline{1-21}
ArcFace FAR $10^{-3}$ ($<0.76$)     & 0.007 & 0.038 & 0.0143 & 0.2737 & 0.0438 & 0.0031 & 0.001 & 0.0279 & 0.0035 & 0.1185 & 0.0085 & 0.0408 & 0.0161 & 0.2699 & 0.0373 & 0.0021 & 0.0003 & 0.0281 & 0.0034 & 0.1168 \\
ArcFace FAR $10^{-4}$ ($<0.69$)     & 0.002 & 0.0114 & 0.0038 & 0.1446 & 0.0146 & 0.0008 & 0.0000 & 0.0058 & 0.0004 & 0.057 & 0.0022 & 0.0118 & 0.0036 & 0.1452 & 0.0138 & 0.0008 & 0.0000 & 0.0036 & 0.0001 & 0.0542 \\
ArcFace FAR $10^{-5}$ ($<0.62$)     & 0.0001 & 0.0007 & 0.0006 & 0.028 & 0.0015 & 0.0001 & 0.0000 & 0.0002 & 0.0000 & 0.0093 & 0.0001 & 0.0011 & 0.0004 & 0.0267 & 0.0014 & 0.0001 & 0.0000 & 0.0004 & 0.0001 & 0.0081 \\
\cdashline{1-21}
AdaFace FAR $10^{-3}$ ($<0.80$)     & 0.0033 & 0.0318 & 0.0076 & 0.359 & 0.0655 & 0.0094 & 0.0026 & 0.0348 & 0.0144 & 0.2063 & 0.0056 & 0.0322 & 0.0084 & 0.3495 & 0.0573 & 0.0073 & 0.0024 & 0.0328 & 0.0127 & 0.2013 \\
AdaFace FAR $10^{-4}$ ($<0.75$)     & 0.0008 & 0.0148 & 0.0035 & 0.2524 & 0.0328 & 0.0044 & 0.0009 & 0.0073 & 0.0062 & 0.1297 & 0.0018 & 0.0168 & 0.003 & 0.252 & 0.0264 & 0.0035 & 0.001 & 0.0059 & 0.0058 & 0.1319 \\
\hline

\end{tabular}
}
\end{center}
\caption{Quantitative experiments on FaceForensics++~\cite{faceforensics++} after tuning the victim system's identity encoder (ArcFace) on learned adversarial examples \cite{ReFace}. We show the detailed results for each attack using AdaFace as the surrogate model.}
\label{t:tuend_adaface_details}
\end{table*}

\begin{table*}[h!]
\begin{center}
\resizebox{1.0\textwidth}{!}{
\begin{tabular}{c|cccccccccc|cccccccccc}
\toprule
&\multicolumn{20}{c}{Attack Type} \\
&\multicolumn{10}{c}{Cosine Distance Objective} &\multicolumn{10}{c}{Euclidean Distance Objective}\\
Evaluation Models             & Snow & Elastic & Fog & Frank Wolfe & Gabor & JPEG L1 & JPEG L2 & JPEG L$\infty$ & PGD L2 & PGD L$\infty$ & Snow & Elastic & Fog & Frank Wolfe & Gabor & JPEG L1 & JPEG L2 & JPEG L$\infty$ & PGD L2 & PGD L$\infty$\\
\cline{1-21}
ArcFace FAR $10^{-3}$ ($<0.76$)     & 0.0071 & 0.03 & 0.0167 & 0.1558 & 0.0082 & 0.0024 & 0.0003 & 0.0172 & 0.0038 & 0.0162 & 0.0073 & 0.0318 & 0.0207 & 0.158 & 0.009 & 0.0029 & 0.0004 & 0.0166 & 0.0034 & 0.0165 \\
ArcFace FAR $10^{-4}$ ($<0.69$)     & 0.0018 & 0.0101 & 0.0044 & 0.0684 & 0.0022 & 0.0007 & 0.0000 & 0.0021 & 0.0004 & 0.0046 & 0.0015 & 0.012 & 0.0068 & 0.062 & 0.0022 & 0.0008 & 0.0000 & 0.0021 & 0.0002 & 0.0049 \\
ArcFace FAR $10^{-5}$ ($<0.62$)     & 0.0001 & 0.0015 & 0.0004 & 0.013 & 0.0000 & 0.0001 & 0.0000 & 0.0000 & 0.0001 & 0.0001 & 0.0000 & 0.001 & 0.0011 & 0.0075 & 0.0001 & 0.0000 & 0.0000 & 0.0000 & 0.0000 & 0.0003 \\
\cdashline{1-21}
AdaFace FAR $10^{-3}$ ($<0.80$)     & 0.0101 & 0.0563 & 0.0245 & 0.2031 & 0.0161 & 0.0107 & 0.0021 & 0.0171 & 0.0158 & 0.0407 & 0.009 & 0.0552 & 0.0285 & 0.1969 & 0.0142 & 0.0121 & 0.003 & 0.0167 & 0.0148 & 0.0392 \\
AdaFace FAR $10^{-4}$ ($<0.75$)     & 0.0037 & 0.0293 & 0.0106 & 0.1198 & 0.0045 & 0.0049 & 0.0003 & 0.0027 & 0.0077 & 0.0172 & 0.0034 & 0.0304 & 0.0122 & 0.1133 & 0.0057 & 0.0067 & 0.0007 & 0.0029 & 0.0068 & 0.0193 \\
\hline

\end{tabular}
}
\end{center}
\caption{Quantitative experiments on FaceForensics++~\cite{faceforensics++} after tuning the victim system's identity encoder (ArcFace) on learned adversarial examples \cite{ReFace}. We show the detailed results for each attack using MagFace as the surrogate model.}
\label{t:tuend_magface_details}
\end{table*}

\begin{table*}[h!]
\begin{center}
\resizebox{1.0\textwidth}{!}{
\begin{tabular}{c|cccccccccc|cccccccccc}
\toprule
&\multicolumn{20}{c}{Attack Type} \\
&\multicolumn{10}{c}{Cosine Distance Objective} &\multicolumn{10}{c}{Euclidean Distance Objective}\\
Evaluation Models             & Snow & Elastic & Fog & Frank Wolfe & Gabor & JPEG L1 & JPEG L2 & JPEG L$\infty$ & PGD L2 & PGD L$\infty$ & Snow & Elastic & Fog & Frank Wolfe & Gabor & JPEG L1 & JPEG L2 & JPEG L$\infty$ & PGD L2 & PGD L$\infty$\\
\cline{1-21}
ArcFace FAR $10^{-3}$ ($<0.76$)     & 0.0138 & 0.0451 & 0.0158 & 0.2808 & 0.056 & 0.002 & 0.0011 & 0.0306 & 0.0033 & 0.1199 & 0.0114 & 0.0414 & 0.0197 & 0.2775 & 0.0569 & 0.002 & 0.0015 & 0.0342 & 0.0031 & 0.1147 \\
ArcFace FAR $10^{-4}$ ($<0.69$)     & 0.0046 & 0.0152 & 0.004 & 0.1602 & 0.0216 & 0.0002 & 0.0003 & 0.0071 & 0.0003 & 0.0551 & 0.0042 & 0.016 & 0.0053 & 0.1571 & 0.0221 & 0.0002 & 0.0004 & 0.0075 & 0.0003 & 0.0535 \\
ArcFace FAR $10^{-5}$ ($<0.62$)     & 0.0004 & 0.0009 & 0.0001 & 0.0294 & 0.0028 & 0.0000 & 0.0001 & 0.0007 & 0.0001 & 0.0112 & 0.0006 & 0.0014 & 0.0005 & 0.0324 & 0.0022 & 0.0000 & 0.0001 & 0.0006 & 0.0000 & 0.0104 \\
\cdashline{1-21}
AdaFace FAR $10^{-3}$ ($<0.80$)     & 0.0125 & 0.0838 & 0.0188 & 0.373 & 0.0908 & 0.0116 & 0.0034 & 0.0354 & 0.0152 & 0.1994 & 0.0143 & 0.0842 & 0.0213 & 0.3694 & 0.0851 & 0.009 & 0.0025 & 0.038 & 0.0142 & 0.1992 \\
AdaFace FAR $10^{-4}$ ($<0.75$)     & 0.0053 & 0.0485 & 0.0072 & 0.2791 & 0.0431 & 0.0066 & 0.0021 & 0.0096 & 0.0066 & 0.1302 & 0.0068 & 0.0479 & 0.0099 & 0.2798 & 0.0441 & 0.0042 & 0.0014 & 0.0091 & 0.0062 & 0.1293 \\
\hline

\end{tabular}
}
\end{center}
\caption{Quantitative experiments on FaceForensics++~\cite{faceforensics++} after tuning the victim system's identity encoder (ArcFace) on learned adversarial examples \cite{ReFace}. We show the detailed results for each attack using ElasticFace as the surrogate model.}
\label{t:tuend_elastiface_details}
\end{table*}

\begin{table*}[h!]
\begin{center}
\resizebox{1.0\textwidth}{!}{
\begin{tabular}{c|cccccccccc|cccccccccc}
\toprule
&\multicolumn{20}{c}{Attack Type} \\
&\multicolumn{10}{c}{Cosine Distance Objective} &\multicolumn{10}{c}{Euclidean Distance Objective}\\
Evaluation Models             & Snow & Elastic & Fog & Frank Wolfe & Gabor & JPEG L1 & JPEG L2 & JPEG L$\infty$ & PGD L2 & PGD L$\infty$ & Snow & Elastic & Fog & Frank Wolfe & Gabor & JPEG L1 & JPEG L2 & JPEG L$\infty$ & PGD L2 & PGD L$\infty$\\
\cline{1-21}
ArcFace FAR $10^{-3}$ ($<0.76$)     & 0.0148 & 0.0476 & 0.018 & 0.202 & 0.0858 & 0.0051 & 0.0006 & 0.0328 & 0.0037 & 0.0785 & 0.0164 & 0.0473 & 0.0174 & 0.2002 & 0.0854 & 0.0049 & 0.0003 & 0.0329 & 0.0033 & 0.0745 \\
ArcFace FAR $10^{-4}$ ($<0.69$)     & 0.0037 & 0.0141 & 0.0046 & 0.0779 & 0.0381 & 0.0009 & 0.0000 & 0.0061 & 0.0001 & 0.0312 & 0.0033 & 0.0152 & 0.0055 & 0.077 & 0.0354 & 0.0009 & 0.0001 & 0.0066 & 0.0002 & 0.0295 \\
ArcFace FAR $10^{-5}$ ($<0.62$)     & 0.0006 & 0.0006 & 0.0002 & 0.0085 & 0.0061 & 0.0002 & 0.0000 & 0.0005 & 0.0001 & 0.0042 & 0.0003 & 0.001 & 0.0001 & 0.0091 & 0.0054 & 0.0002 & 0.0000 & 0.0007 & 0.0001 & 0.004 \\
\cdashline{1-21}
AdaFace FAR $10^{-3}$ ($<0.80$)     & 0.0157 & 0.0974 & 0.0204 & 0.2485 & 0.1255 & 0.0138 & 0.0028 & 0.0321 & 0.0134 & 0.1515 & 0.0151 & 0.1029 & 0.0218 & 0.2467 & 0.1247 & 0.0126 & 0.0024 & 0.0351 & 0.0146 & 0.1546 \\
AdaFace FAR $10^{-4}$ ($<0.75$)     & 0.0064 & 0.0502 & 0.0081 & 0.1429 & 0.068 & 0.0069 & 0.0012 & 0.0078 & 0.0068 & 0.0846 & 0.0056 & 0.0516 & 0.0081 & 0.1385 & 0.0665 & 0.0076 & 0.0004 & 0.0067 & 0.0068 & 0.0893 \\
\hline

\end{tabular}
}
\end{center}
\caption{Quantitative experiments on FaceForensics++~\cite{faceforensics++} after tuning the victim system's identity encoder (ArcFace) on learned adversarial examples \cite{ReFace}. We show the detailed results for each attack using CosFace as the surrogate model.}
\label{t:tuend_cosface_details}
\end{table*}

\begin{table*}[h!]
\begin{center}
\resizebox{1.0\textwidth}{!}{
\begin{tabular}{c|cccccccccc|cccccccccc}
\toprule
&\multicolumn{20}{c}{Attack Type} \\
&\multicolumn{10}{c}{Cosine Distance Objective} &\multicolumn{10}{c}{Euclidean Distance Objective}\\
Evaluation Models             & Snow & Elastic & Fog & Frank Wolfe & Gabor & JPEG L1 & JPEG L2 & JPEG L$\infty$ & PGD L2 & PGD L$\infty$ & Snow & Elastic & Fog & Frank Wolfe & Gabor & JPEG L1 & JPEG L2 & JPEG L$\infty$ & PGD L2 & PGD L$\infty$\\
\cline{1-21}
ArcFace FAR $10^{-3}$ ($<0.76$)     & 0.0021 & 0.0212 & 0.0051 & 0.0046 & 0.0009 & 0.0016 & 0.0013 & 0.028 & 0.0039 & 0.0008 & 0.0014 & 0.0167 & 0.0055 & 0.0059 & 0.0009 & 0.0019 & 0.0011 & 0.0284 & 0.0027 & 0.0007 \\
ArcFace FAR $10^{-4}$ ($<0.69$)     & 0.0005 & 0.0078 & 0.0011 & 0.0018 & 0.0002 & 0.0001 & 0.0004 & 0.0048 & 0.0002 & 0.0000 & 0.0003 & 0.0057 & 0.0012 & 0.0023 & 0.0000 & 0.0001 & 0.0001 & 0.0069 & 0.0002 & 0.0001 \\
ArcFace FAR $10^{-5}$ ($<0.62$)     & 0.0000 & 0.0007 & 0.0000 & 0.0002 & 0.0000 & 0.0000 & 0.0000 & 0.0002 & 0.0001 & 0.0000 & 0.0000 & 0.0007 & 0.0001 & 0.0005 & 0.0000 & 0.0000 & 0.0001 & 0.0002 & 0.0000 & 0.0 \\
\cdashline{1-21}
AdaFace FAR $10^{-3}$ ($<0.80$)     & 0.0023 & 0.0315 & 0.0057 & 0.0116 & 0.0025 & 0.0089 & 0.004 & 0.0294 & 0.0146 & 0.0054 & 0.0015 & 0.031 & 0.0054 & 0.0103 & 0.0033 & 0.0095 & 0.0024 & 0.0288 & 0.014 & 0.004 \\
AdaFace FAR $10^{-4}$ ($<0.75$)     & 0.0016 & 0.0179 & 0.0015 & 0.0029 & 0.0003 & 0.004 & 0.0018 & 0.0078 & 0.006 & 0.0022 & 0.0002 & 0.0158 & 0.0016 & 0.0047 & 0.0004 & 0.0042 & 0.0004 & 0.0069 & 0.007 & 0.0012 \\
\hline

\end{tabular}
}
\end{center}
\caption{Quantitative experiments on FaceForensics++~\cite{faceforensics++} after tuning the victim system's identity encoder (ArcFace) on learned adversarial examples \cite{ReFace}. We show the detailed results for each attack using FaceNet as the surrogate model.}
\label{t:tuend_facenet_details}
\end{table*}

\begin{table*}[h!]
\begin{center}
\resizebox{1.0\textwidth}{!}{
\begin{tabular}{c|cccccccccc|cccccccccc}
\toprule
&\multicolumn{20}{c}{Attack Type} \\
&\multicolumn{10}{c}{Cosine Distance Objective} &\multicolumn{10}{c}{Euclidean Distance Objective}\\
Evaluation Models             & Snow & Elastic & Fog & Frank Wolfe & Gabor & JPEG L1 & JPEG L2 & JPEG L$\infty$ & PGD L2 & PGD L$\infty$ & Snow & Elastic & Fog & Frank Wolfe & Gabor & JPEG L1 & JPEG L2 & JPEG L$\infty$ & PGD L2 & PGD L$\infty$\\
\cline{1-21}
ArcFace FAR $10^{-3}$ ($<0.76$)     & 0.0185 & 0.0685 & 0.0253 & 0.6202 & 0.1567 & 0.0039 & 0.0005 & 0.0305 & 0.0032 & 0.2062 & 0.0191 & 0.0744 & 0.0256 & 0.6171 & 0.1585 & 0.0041 & 0.0006 & 0.0306 & 0.0036 & 0.2114 \\
ArcFace FAR $10^{-4}$ ($<0.69$)     & 0.006 & 0.0251 & 0.0067 & 0.4887 & 0.0804 & 0.0011 & 0.0000 & 0.0079 & 0.0003 & 0.1166 & 0.0072 & 0.0247 & 0.0061 & 0.4842 & 0.0805 & 0.0011 & 0.0001 & 0.006 & 0.0002 & 0.1199 \\
ArcFace FAR $10^{-5}$ ($<0.62$)     & 0.0007 & 0.0029 & 0.0009 & 0.1935 & 0.0178 & 0.0003 & 0.0000 & 0.0004 & 0.0000 & 0.0276 & 0.0011 & 0.0018 & 0.0008 & 0.1861 & 0.0156 & 0.0002 & 0.0000 & 0.0007 & 0.0000 & 0.0259 \\
\cdashline{1-21}
AdaFace FAR $10^{-3}$ ($<0.80$)     & 0.0458 & 0.2527 & 0.0613 & 0.6708 & 0.2144 & 0.0097 & 0.0023 & 0.0301 & 0.0147 & 0.2858 & 0.0475 & 0.2587 & 0.0569 & 0.6818 & 0.2256 & 0.013 & 0.0018 & 0.0298 & 0.0136 & 0.2936 \\
AdaFace FAR $10^{-4}$ ($<0.75$)     & 0.0214 & 0.1562 & 0.0293 & 0.5855 & 0.1361 & 0.0055 & 0.0008 & 0.0072 & 0.0067 & 0.2012 & 0.022 & 0.1719 & 0.0277 & 0.6014 & 0.1391 & 0.0062 & 0.0008 & 0.0075 & 0.0062 & 0.2031 \\
\hline

\end{tabular}
}
\end{center}
\caption{Quantitative experiments on FaceForensics++~\cite{faceforensics++} after tuning the victim system's identity encoder (ArcFace) on learned adversarial examples \cite{ReFace}. We show the detailed results for each attack using ArcFace as the surrogate model.}
\label{t:tuend_arcface_details}
\end{table*}

\begin{table*}[h!]
\begin{center}
\resizebox{1.0\textwidth}{!}{
\begin{tabular}{c|cccccccccc|cccccccccc}
\toprule
&\multicolumn{20}{c}{Attack Type} \\
&\multicolumn{10}{c}{Cosine Distance Objective} &\multicolumn{10}{c}{Euclidean Distance Objective}\\
Evaluation Models             & Snow & Elastic & Fog & Frank Wolfe & Gabor & JPEG L1 & JPEG L2 & JPEG L$\infty$ & PGD L2 & PGD L$\infty$ & Snow & Elastic & Fog & Frank Wolfe & Gabor & JPEG L1 & JPEG L2 & JPEG L$\infty$ & PGD L2 & PGD L$\infty$\\
\cline{1-21}
ArcFace FAR $10^{-3}$ ($<0.76$)     & 0.0022 & 0.0053 & 0.0032 & 0.0033 & 0.0006 & 0.003 & 0.0005 & 0.001 & 0.0055 & 0.0023 & 0.0019 & 0.0041 & 0.0046 & 0.0033 & 0.0008 & 0.0024 & 0.0004 & 0.0015 & 0.0048 & 0.0033 \\
ArcFace FAR $10^{-4}$ ($<0.69$)     & 0.0002 & 0.0024 & 0.001 & 0.0006 & 0.0002 & 0.0008 & 0.0002 & 0.0001 & 0.0015 & 0.0007 & 0.0001 & 0.0009 & 0.0012 & 0.0011 & 0.0002 & 0.0006 & 0.0001 & 0.0000 & 0.0013 & 0.0009 \\
ArcFace FAR $10^{-5}$ ($<0.62$)     & 0.0000 & 0.0003 & 0.0000 & 0.0001 & 0.0000 & 0.0000 & 0.0000 & 0.0000 & 0.0001 & 0.0000 & 0.0000 & 0.0000 & 0.0000 & 0.0000 & 0.0000 & 0.0000 & 0.0001 & 0.0000 & 0.0000 & 0.0001 \\
\cdashline{1-21}
AdaFace FAR $10^{-3}$ ($<0.80$)     & 0.0036 & 0.0146 & 0.0059 & 0.0091 & 0.0022 & 0.0096 & 0.0023 & 0.0018 & 0.0156 & 0.0101 & 0.0038 & 0.0146 & 0.0075 & 0.0079 & 0.0011 & 0.0094 & 0.0018 & 0.0025 & 0.0146 & 0.0109 \\
AdaFace FAR $10^{-4}$ ($<0.75$)     & 0.0015 & 0.0077 & 0.002 & 0.0037 & 0.0006 & 0.0036 & 0.0007 & 0.0003 & 0.0078 & 0.0044 & 0.0012 & 0.0081 & 0.0023 & 0.0026 & 0.0002 & 0.0046 & 0.0002 & 0.0001 & 0.0064 & 0.0038 \\
\hline

\end{tabular}
}
\end{center}
\caption{Quantitative experiments on FaceForensics++~\cite{faceforensics++} after tuning the victim system's identity encoder (ArcFace) on learned adversarial examples \cite{ReFace} and distortion attacks \cite{OpenAIRobust}. We show the detailed results for each attack using ResNet50 trained on ImageNet as the surrogate model.}
\label{t:dis_tuned_resnet_details}
\end{table*}

\begin{table*}[h!]
\begin{center}
\resizebox{1.0\textwidth}{!}{
\begin{tabular}{c|cccccccccc|cccccccccc}
\toprule
&\multicolumn{20}{c}{Attack Type} \\
&\multicolumn{10}{c}{Cosine Distance Objective} &\multicolumn{10}{c}{Euclidean Distance Objective}\\
Evaluation Models             & Snow & Elastic & Fog & Frank Wolfe & Gabor & JPEG L1 & JPEG L2 & JPEG L$\infty$ & PGD L2 & PGD L$\infty$ & Snow & Elastic & Fog & Frank Wolfe & Gabor & JPEG L1 & JPEG L2 & JPEG L$\infty$ & PGD L2 & PGD L$\infty$\\
\cline{1-21}
ArcFace FAR $10^{-3}$ ($<0.76$)     & 0.0067 & 0.0351 & 0.0118 & 0.3165 & 0.0229 & 0.0027 & 0.0007 & 0.0007 & 0.0036 & 0.1667 & 0.0081 & 0.0361 & 0.0123 & 0.3198 & 0.0246 & 0.0032 & 0.0006 & 0.0002 & 0.0048 & 0.1554 \\
ArcFace FAR $10^{-4}$ ($<0.69$)     & 0.0015 & 0.0085 & 0.0034 & 0.1945 & 0.0051 & 0.0005 & 0.0001 & 0.0001 & 0.0014 & 0.0855 & 0.0019 & 0.0093 & 0.0024 & 0.1924 & 0.0065 & 0.0008 & 0.0001 & 0.0000 & 0.0019 & 0.08 \\
ArcFace FAR $10^{-5}$ ($<0.62$)     & 0.0001 & 0.0005 & 0.0005 & 0.0468 & 0.0003 & 0.0000 & 0.0000 & 0.0000 & 0.0000 & 0.0159 & 0.0002 & 0.0009 & 0.0000 & 0.0486 & 0.0009 & 0.0000 & 0.0000 & 0.0000 & 0.0001 & 0.0154 \\
\cdashline{1-21}
AdaFace FAR $10^{-3}$ ($<0.80$)     & 0.0035 & 0.0281 & 0.0063 & 0.3891 & 0.0364 & 0.0109 & 0.0024 & 0.0019 & 0.0149 & 0.2634 & 0.0042 & 0.0261 & 0.0077 & 0.3899 & 0.0409 & 0.0101 & 0.0023 & 0.0005 & 0.0176 & 0.2478 \\
AdaFace FAR $10^{-4}$ ($<0.75$)     & 0.0009 & 0.0113 & 0.0022 & 0.298 & 0.0151 & 0.0046 & 0.0002 & 0.0003 & 0.007 & 0.174 & 0.001 & 0.0123 & 0.0023 & 0.3048 & 0.0165 & 0.0041 & 0.0004 & 0.0000 & 0.0085 & 0.1654 \\
\hline

\end{tabular}
}
\end{center}
\caption{Quantitative experiments on FaceForensics++~\cite{faceforensics++} after tuning the victim system's identity encoder (ArcFace) on learned adversarial examples \cite{ReFace} and distortion attacks \cite{OpenAIRobust}. We show the detailed results for each attack using AdaFace as the surrogate model.}
\label{t:dis_tuned_adaface_details}
\end{table*}

\begin{table*}[h!]
\begin{center}
\resizebox{1.0\textwidth}{!}{
\begin{tabular}{c|cccccccccc|cccccccccc}
\toprule
&\multicolumn{20}{c}{Attack Type} \\
&\multicolumn{10}{c}{Cosine Distance Objective} &\multicolumn{10}{c}{Euclidean Distance Objective}\\
Evaluation Models             & Snow & Elastic & Fog & Frank Wolfe & Gabor & JPEG L1 & JPEG L2 & JPEG L$\infty$ & PGD L2 & PGD L$\infty$ & Snow & Elastic & Fog & Frank Wolfe & Gabor & JPEG L1 & JPEG L2 & JPEG L$\infty$ & PGD L2 & PGD L$\infty$\\
\cline{1-21}
ArcFace FAR $10^{-3}$ ($<0.76$)     & 0.0068 & 0.0272 & 0.0136 & 0.0957 & 0.0064 & 0.0027 & 0.0007 & 0.0059 & 0.0039 & 0.0215 & 0.0088 & 0.0281 & 0.0165 & 0.0974 & 0.0076 & 0.0034 & 0.0005 & 0.0069 & 0.0032 & 0.017 \\
ArcFace FAR $10^{-4}$ ($<0.69$)     & 0.0021 & 0.0077 & 0.0027 & 0.0475 & 0.0011 & 0.0008 & 0.0003 & 0.0013 & 0.0015 & 0.0071 & 0.0018 & 0.0103 & 0.0045 & 0.0443 & 0.0009 & 0.0008 & 0.0001 & 0.0026 & 0.0002 & 0.0061 \\
ArcFace FAR $10^{-5}$ ($<0.62$)     & 0.0002 & 0.0001 & 0.0001 & 0.0111 & 0.0000 & 0.0001 & 0.0000 & 0.0001 & 0.0001 & 0.0006 & 0.0000 & 0.0005 & 0.0004 & 0.009 & 0.0001 & 0.0001 & 0.0000 & 0.0007 & 0.0000 & 0.0009 \\
\cdashline{1-21}
AdaFace FAR $10^{-3}$ ($<0.80$)     & 0.0092 & 0.05 & 0.018 & 0.1323 & 0.0126 & 0.0089 & 0.002 & 0.0072 & 0.0145 & 0.0485 & 0.0088 & 0.0496 & 0.0222 & 0.136 & 0.0121 & 0.0088 & 0.0015 & 0.0097 & 0.0131 & 0.0387 \\
AdaFace FAR $10^{-4}$ ($<0.75$)     & 0.0028 & 0.0248 & 0.0084 & 0.0863 & 0.0044 & 0.0038 & 0.0005 & 0.0018 & 0.007 & 0.0262 & 0.0034 & 0.0276 & 0.0095 & 0.09 & 0.0044 & 0.0044 & 0.0007 & 0.0037 & 0.0056 & 0.0192 \\
\hline

\end{tabular}
}
\end{center}
\caption{Quantitative experiments on FaceForensics++~\cite{faceforensics++} after tuning the victim system's identity encoder (ArcFace) on learned adversarial examples \cite{ReFace} and distortion attacks \cite{OpenAIRobust}. We show the detailed results for each attack using MagFace as the surrogate model.}
\label{t:dis_tuned_magface_details}
\end{table*}

\begin{table*}[h!]
\begin{center}
\resizebox{1.0\textwidth}{!}{
\begin{tabular}{c|cccccccccc|cccccccccc}
\toprule
&\multicolumn{20}{c}{Attack Type} \\
&\multicolumn{10}{c}{Cosine Distance Objective} &\multicolumn{10}{c}{Euclidean Distance Objective}\\
Evaluation Models             & Snow & Elastic & Fog & Frank Wolfe & Gabor & JPEG L1 & JPEG L2 & JPEG L$\infty$ & PGD L2 & PGD L$\infty$ & Snow & Elastic & Fog & Frank Wolfe & Gabor & JPEG L1 & JPEG L2 & JPEG L$\infty$ & PGD L2 & PGD L$\infty$\\
\cline{1-21}
ArcFace FAR $10^{-3}$ ($<0.76$)     & 0.0107 & 0.0399 & 0.0128 & 0.333 & 0.0333 & 0.0032 & 0.0013 & 0.0008 & 0.0039 & 0.2002 & 0.0105 & 0.0357 & 0.0127 & 0.3387 & 0.034 & 0.0058 & 0.0008 & 0.0008 & 0.0031 & 0.2045 \\
ArcFace FAR $10^{-4}$ ($<0.69$)     & 0.0035 & 0.0128 & 0.0034 & 0.209 & 0.0109 & 0.0012 & 0.0006 & 0.0000 & 0.0006 & 0.1085 & 0.0035 & 0.0116 & 0.0036 & 0.2093 & 0.0128 & 0.0023 & 0.0001 & 0.0003 & 0.0006 & 0.1094 \\
ArcFace FAR $10^{-5}$ ($<0.62$)     & 0.0002 & 0.0004 & 0.0001 & 0.0521 & 0.001 & 0.0000 & 0.0000 & 0.0000 & 0.0001 & 0.0272 & 0.0003 & 0.0015 & 0.0003 & 0.0526 & 0.0013 & 0.0000 & 0.0000 & 0.0000 & 0.0001 & 0.0247 \\
\cdashline{1-21}
AdaFace FAR $10^{-3}$ ($<0.80$)     & 0.0114 & 0.0757 & 0.0158 & 0.4273 & 0.0594 & 0.0111 & 0.0029 & 0.0006 & 0.0142 & 0.2984 & 0.0123 & 0.0768 & 0.0154 & 0.4232 & 0.0571 & 0.0158 & 0.0016 & 0.0011 & 0.0149 & 0.2996 \\
AdaFace FAR $10^{-4}$ ($<0.75$)     & 0.0058 & 0.0416 & 0.0063 & 0.3345 & 0.0261 & 0.0059 & 0.0008 & 0.0000 & 0.0064 & 0.2142 & 0.0063 & 0.0391 & 0.0064 & 0.3329 & 0.0255 & 0.009 & 0.0004 & 0.0002 & 0.0061 & 0.2127 \\
\hline

\end{tabular}
}
\end{center}
\caption{Quantitative experiments on FaceForensics++~\cite{faceforensics++} after tuning the victim system's identity encoder (ArcFace) on learned adversarial examples \cite{ReFace} and distortion attacks \cite{OpenAIRobust}. We show the detailed results for each attack using ElasticFace as the surrogate model.}
\label{t:dis_tuned_elasticface_details}
\end{table*}

\begin{table*}[h!]
\begin{center}
\resizebox{1.0\textwidth}{!}{
\begin{tabular}{c|cccccccccc|cccccccccc}
\toprule
&\multicolumn{20}{c}{Attack Type} \\
&\multicolumn{10}{c}{Cosine Distance Objective} &\multicolumn{10}{c}{Euclidean Distance Objective}\\
Evaluation Models             & Snow & Elastic & Fog & Frank Wolfe & Gabor & JPEG L1 & JPEG L2 & JPEG L$\infty$ & PGD L2 & PGD L$\infty$ & Snow & Elastic & Fog & Frank Wolfe & Gabor & JPEG L1 & JPEG L2 & JPEG L$\infty$ & PGD L2 & PGD L$\infty$\\
\cline{1-21}
ArcFace FAR $10^{-3}$ ($<0.76$)     & 0.0147 & 0.0434 & 0.0114 & 0.1734 & 0.0519 & 0.0036 & 0.0014 & 0.0007 & 0.0044 & 0.1706 & 0.0144 & 0.0447 & 0.0157 & 0.1697 & 0.0535 & 0.0045 & 0.0006 & 0.0006 & 0.0048 & 0.1589 \\
ArcFace FAR $10^{-4}$ ($<0.69$)     & 0.004 & 0.0106 & 0.0031 & 0.0857 & 0.0154 & 0.0011 & 0.0005 & 0.0001 & 0.0014 & 0.0884 & 0.0034 & 0.0127 & 0.0043 & 0.0848 & 0.0203 & 0.0012 & 0.0003 & 0.0002 & 0.0015 & 0.0797 \\
ArcFace FAR $10^{-5}$ ($<0.62$)     & 0.0006 & 0.0004 & 0.0001 & 0.0172 & 0.0018 & 0.0000 & 0.0000 & 0.0000 & 0.0002 & 0.0196 & 0.0001 & 0.0008 & 0.0003 & 0.0168 & 0.003 & 0.0001 & 0.0000 & 0.0000 & 0.0001 & 0.0181 \\
\cdashline{1-21}
AdaFace FAR $10^{-3}$ ($<0.80$)     & 0.0136 & 0.0823 & 0.0153 & 0.2371 & 0.0886 & 0.0112 & 0.0025 & 0.001 & 0.0187 & 0.2657 & 0.0125 & 0.0911 & 0.0184 & 0.231 & 0.0855 & 0.0105 & 0.002 & 0.0003 & 0.0165 & 0.2568 \\
AdaFace FAR $10^{-4}$ ($<0.75$)     & 0.0052 & 0.0395 & 0.0056 & 0.1612 & 0.0418 & 0.0049 & 0.001 & 0.0001 & 0.0082 & 0.1762 & 0.0038 & 0.0471 & 0.007 & 0.1608 & 0.0421 & 0.0059 & 0.0005 & 0.0000 & 0.007 & 0.1646 \\
\hline

\end{tabular}
}
\end{center}
\caption{Quantitative experiments on FaceForensics++~\cite{faceforensics++} after tuning the victim system's identity encoder (ArcFace) on learned adversarial examples \cite{ReFace} and distortion attacks \cite{OpenAIRobust}. We show the detailed results for each attack using CosFace as the surrogate model.}
\label{t:dis_tuned_cosface_details}
\end{table*}

\begin{table*}[h!]
\begin{center}
\resizebox{1.0\textwidth}{!}{
\begin{tabular}{c|cccccccccc|cccccccccc}
\toprule
&\multicolumn{20}{c}{Attack Type} \\
&\multicolumn{10}{c}{Cosine Distance Objective} &\multicolumn{10}{c}{Euclidean Distance Objective}\\
Evaluation Models             & Snow & Elastic & Fog & Frank Wolfe & Gabor & JPEG L1 & JPEG L2 & JPEG L$\infty$ & PGD L2 & PGD L$\infty$ & Snow & Elastic & Fog & Frank Wolfe & Gabor & JPEG L1 & JPEG L2 & JPEG L$\infty$ & PGD L2 & PGD L$\infty$\\
\cline{1-21}
ArcFace FAR $10^{-3}$ ($<0.76$)     & 0.0021 & 0.019 & 0.0041 & 0.003 & 0.0003 & 0.0041 & 0.0007 & 0.0014 & 0.0045 & 0.0013 & 0.0013 & 0.0161 & 0.0031 & 0.0028 & 0.0006 & 0.0043 & 0.0014 & 0.0015 & 0.0032 & 0.0022 \\
ArcFace FAR $10^{-4}$ ($<0.69$)     & 0.0006 & 0.0072 & 0.0008 & 0.0011 & 0.0001 & 0.0012 & 0.0002 & 0.0000 & 0.0015 & 0.0006 & 0.0003 & 0.0061 & 0.0006 & 0.0003 & 0.0000 & 0.0015 & 0.0003 & 0.0000 & 0.0005 & 0.0007 \\
ArcFace FAR $10^{-5}$ ($<0.62$)     & 0.0000 & 0.0005 & 0.0001 & 0.0002 & 0.0000 & 0.0000 & 0.0000 & 0.0000 & 0.0002 & 0.0001 & 0.0000 & 0.0006 & 0.0001 & 0.0000 & 0.0000 & 0.0000 & 0.0001 & 0.0000 & 0.0000 & 0.0 \\
\cdashline{1-21}
AdaFace FAR $10^{-3}$ ($<0.80$)     & 0.0018 & 0.0325 & 0.0059 & 0.0085 & 0.0014 & 0.0146 & 0.002 & 0.0014 & 0.0152 & 0.004 & 0.0021 & 0.0313 & 0.0052 & 0.0069 & 0.0019 & 0.0142 & 0.0021 & 0.0031 & 0.0135 & 0.0048 \\
AdaFace FAR $10^{-4}$ ($<0.75$)     & 0.0004 & 0.0162 & 0.0016 & 0.0022 & 0.0004 & 0.0075 & 0.0008 & 0.0001 & 0.0067 & 0.0015 & 0.0007 & 0.0167 & 0.0018 & 0.0016 & 0.0005 & 0.0079 & 0.0007 & 0.0003 & 0.0056 & 0.0013 \\
\hline

\end{tabular}
}
\end{center}
\caption{Quantitative experiments on FaceForensics++~\cite{faceforensics++} after tuning the victim system's identity encoder (ArcFace) on learned adversarial examples \cite{ReFace} and distortion attacks \cite{OpenAIRobust}. We show the detailed results for each attack using FaceNet as the surrogate model.}
\label{t:dis_tuned_facenet_details}
\end{table*}

\begin{table*}[h!]
\begin{center}
\resizebox{1.0\textwidth}{!}{
\begin{tabular}{c|cccccccccc|cccccccccc}
\toprule
&\multicolumn{20}{c}{Attack Type} \\
&\multicolumn{10}{c}{Cosine Distance Objective} &\multicolumn{10}{c}{Euclidean Distance Objective}\\
Evaluation Models             & Snow & Elastic & Fog & Frank Wolfe & Gabor & JPEG L1 & JPEG L2 & JPEG L$\infty$ & PGD L2 & PGD L$\infty$ & Snow & Elastic & Fog & Frank Wolfe & Gabor & JPEG L1 & JPEG L2 & JPEG L$\infty$ & PGD L2 & PGD L$\infty$\\
\cline{1-21}
ArcFace FAR $10^{-3}$ ($<0.76$)     & 0.0136 & 0.0387 & 0.0158 & 0.5558 & 0.0738 & 0.0052 & 0.0001 & 0.0008 & 0.0054 & 0.3059 & 0.0138 & 0.0435 & 0.016 & 0.5708 & 0.0678 & 0.0036 & 0.0008 & 0.0006 & 0.0063 & 0.3039 \\
ArcFace FAR $10^{-4}$ ($<0.69$)     & 0.0033 & 0.0125 & 0.0039 & 0.4128 & 0.0267 & 0.0025 & 0.0000 & 0.0005 & 0.002 & 0.1927 & 0.0041 & 0.0141 & 0.0036 & 0.4323 & 0.0271 & 0.0011 & 0.0000 & 0.0000 & 0.0026 & 0.1832 \\
ArcFace FAR $10^{-5}$ ($<0.62$)     & 0.0002 & 0.0009 & 0.0003 & 0.1449 & 0.0052 & 0.0000 & 0.0000 & 0.0000 & 0.0000 & 0.0486 & 0.0006 & 0.001 & 0.0007 & 0.161 & 0.0048 & 0.0000 & 0.0000 & 0.0000 & 0.0001 & 0.0471 \\
\cdashline{1-21}
AdaFace FAR $10^{-3}$ ($<0.80$)     & 0.0306 & 0.1669 & 0.0398 & 0.6178 & 0.107 & 0.0143 & 0.0016 & 0.0016 & 0.016 & 0.386 & 0.0328 & 0.1593 & 0.0391 & 0.6246 & 0.1115 & 0.0126 & 0.0024 & 0.0007 & 0.0173 & 0.3909 \\
AdaFace FAR $10^{-4}$ ($<0.75$)     & 0.0154 & 0.0984 & 0.017 & 0.5221 & 0.0548 & 0.0075 & 0.0005 & 0.0008 & 0.0073 & 0.2887 & 0.0125 & 0.0953 & 0.0162 & 0.5352 & 0.0569 & 0.0067 & 0.0003 & 0.0001 & 0.0072 & 0.2875 \\
\hline

\end{tabular}
}
\end{center}
\caption{Quantitative experiments on FaceForensics++~\cite{faceforensics++} after tuning the victim system's identity encoder (ArcFace) on learned adversarial examples \cite{ReFace} and distortion attacks \cite{OpenAIRobust}. We show the detailed results for each attack using ArcFace as the surrogate model.}
\label{t:dis_tuned_arcface_details}
\end{table*}
\section*{Acknowledgment}
Co-funded by Vinnova Advanced Digitalization, Cyber Security for Industrial Advanced Digitalization. Project grant no. 2023-02996.
\\
Co-funded by the European Union. Views and opinions expressed are however those of the author(s) only and do not necessarily reflect those of the European Union or European Climate, Infrastructure and Environment Executive Agency (CINEA). Neither the European Union nor the granting authority can be held responsible for them. Project grant no. 101069576.

\section*{Disclosure Statement}
Conflict of Interest Statement: The authors declare that the authors has a co-authorship relationship with Vitomir Štruc, who is currently serving as a Guest Editor for IEEE Transactions on Biometrics, Identity and Security (TBIOM) Special Issue on “Generative AI and Large Vision-Language Models for Biometrics”. This co-authorship is on a separate manuscript submitted elsewhere. To ensure a fair and unbiased review process, we request that this manuscript be handled by editors who have no professional or personal connection with any of the authors.

\ifCLASSOPTIONcaptionsoff
  \newpage
\fi



%
\newpage
\clearpage
{\small
\bibliographystyle{Transactions-Bibliography/IEEEtran.bst}
\bibliography{egbib}
}

%

\begin{IEEEbiography}[{\includegraphics[width=1in,height=1.25in,clip,keepaspectratio]{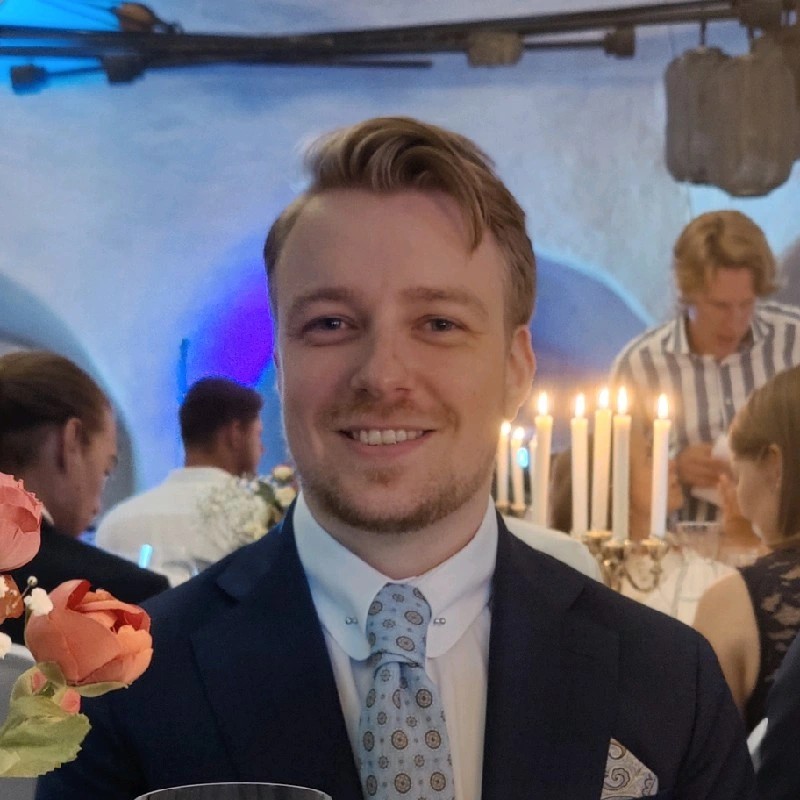}}]{Felix Rosberg}
recieved his M.Sc. in computer science and intelligent systems and Ph.D. in signal processing from Halmstad University, Sweden, in 2020 and 2025, respectively. His research interests include biometrics, generative AI, self-supervised learning, and computer vision for enabling autonomous driving.
\end{IEEEbiography}

\begin{IEEEbiography}[{\includegraphics[width=1in,height=1.25in,clip,keepaspectratio]{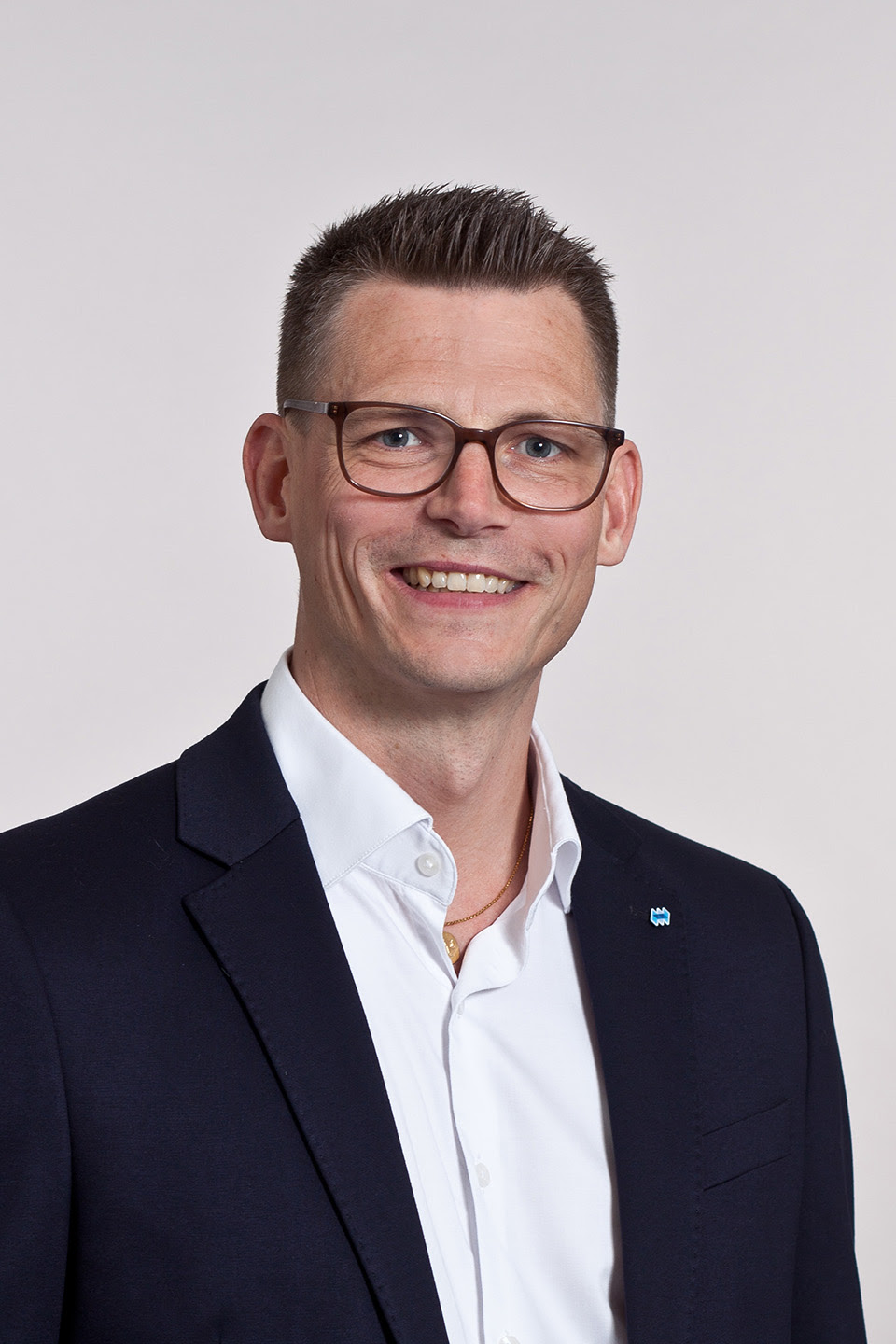}}]{Cristofer Englund}
is a docent and Professor at Halmstad University, Sweden, where he also received his B.Sc. and M.Sc. degrees in electrical engineering and computer science in 2021 and 2023, respectively. He received his PhD in electrical engineering from Chalmers University of Technology, Sweden, in 2007. After several years in the industry and research institute domain, he is currently dean of the School of Information Technology at Halmstad University. His research interests include trustworthy and explainable AI and, computer vision and machine learning for traffic safety applications.
\end{IEEEbiography}

\begin{IEEEbiography}[{\includegraphics[width=1in,height=1.25in,clip,keepaspectratio]{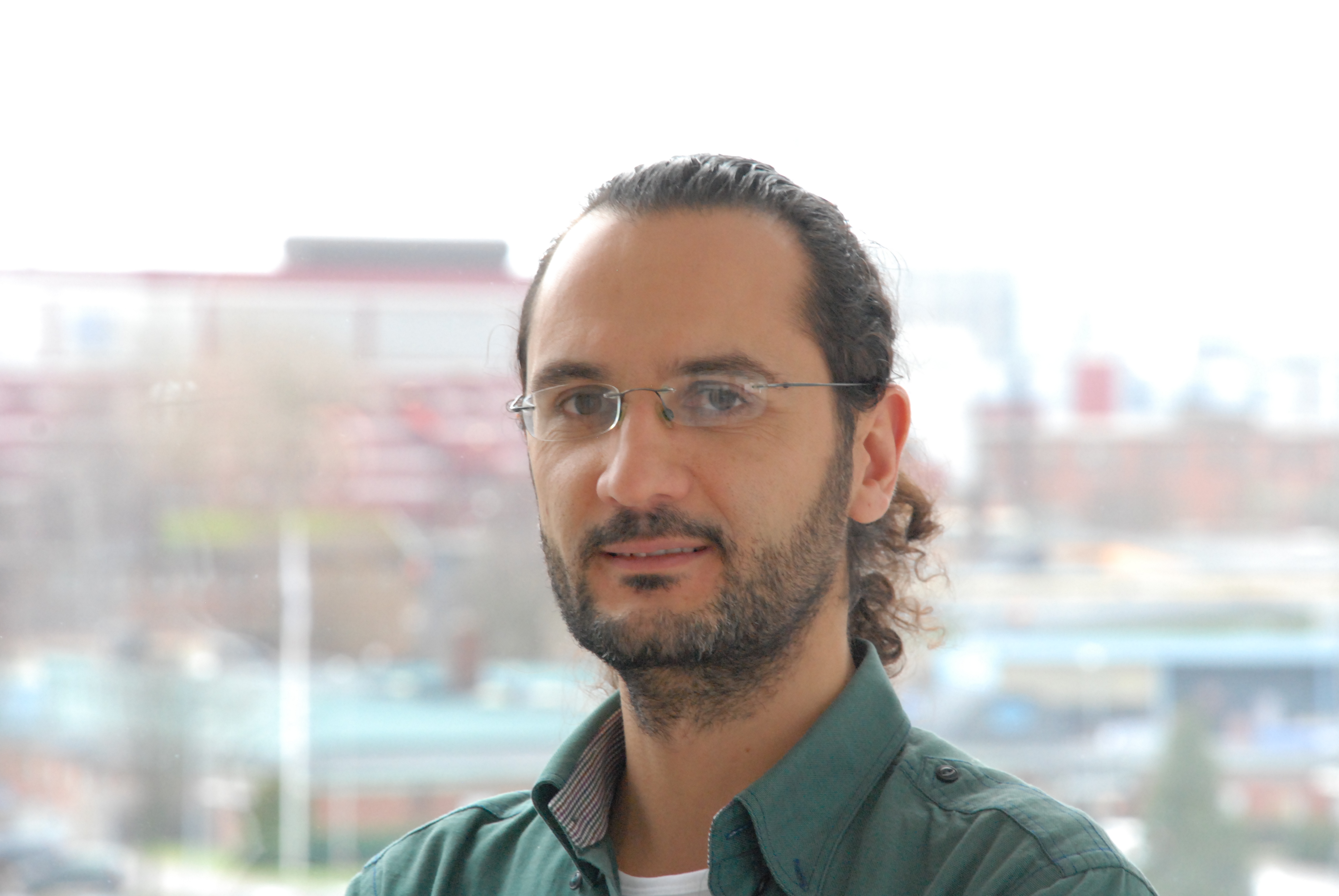}}]{Eren Erdal Aksoy}
received his M.Sc. degree in Mechatronics from the University of Siegen in Germany in 2008. He earned his Ph.D. degree in computer science from the University of Göttingen in Germany in 2012. He is currently employed as an Associate Professor at Halmstad University in Sweden. His research interests include semantic scene perception, computer vision, and robotics.
\end{IEEEbiography}

\begin{IEEEbiography}[{\includegraphics[width=1in,height=1.25in,clip,keepaspectratio]{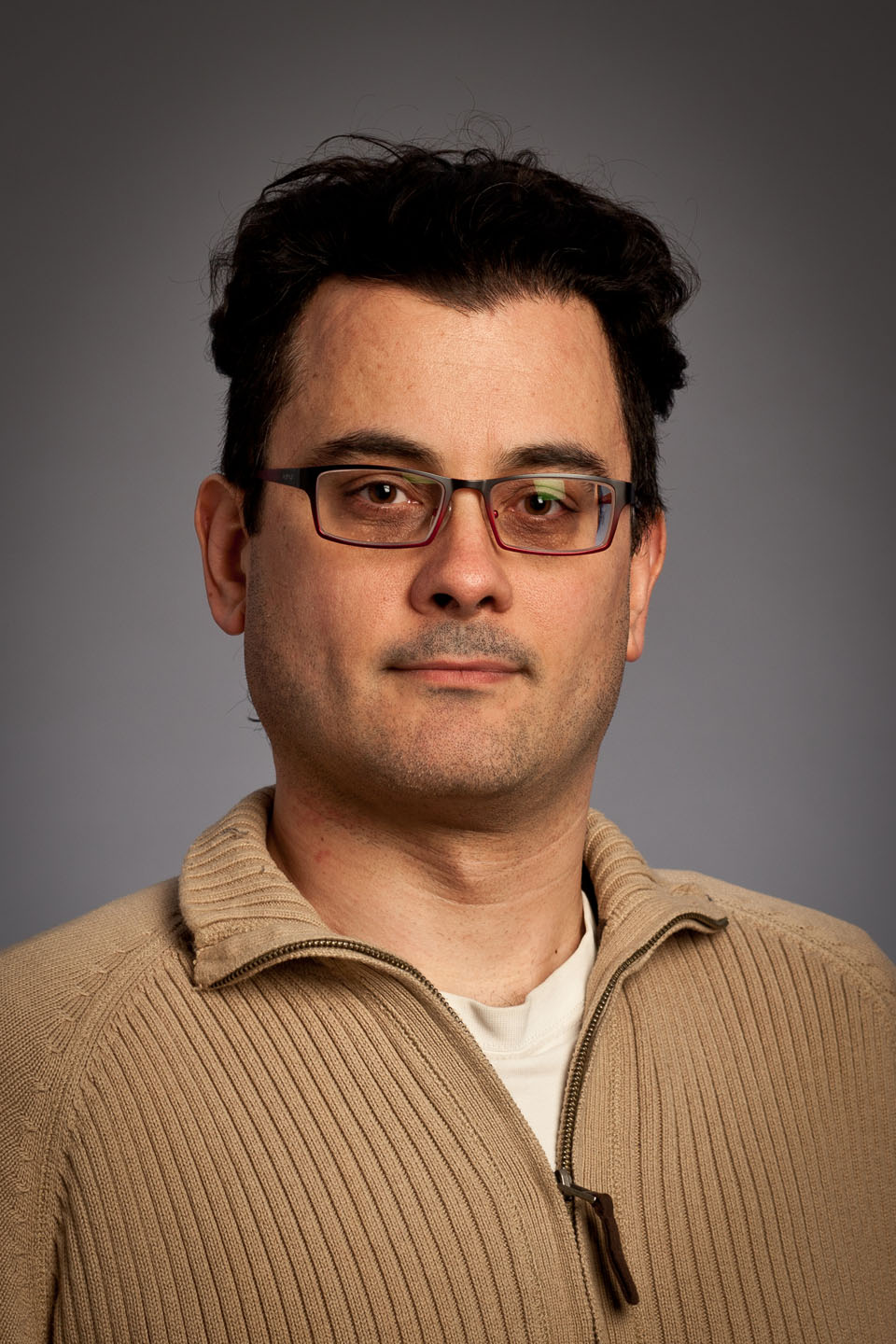}}]{Fernando Alonso-Fernandez}
is a docent and a Professor at Halmstad University, Sweden. He received the M.S./Ph.D. degrees in telecommunications from Universidad Politecnica de Madrid, Spain, in 2003/2008. Since 2010, he has been with Halmstad University, Sweden. His research interests include biometrics and computer vision for security applications.
\end{IEEEbiography}





\end{document}